\documentclass[conference]{IEEEtran}
\IEEEoverridecommandlockouts

\usepackage{fancyhdr}
\usepackage{lipsum} 

\fancypagestyle{plain}{
    \fancyhf{} 
    \fancyfoot[C]{\footnotesize © 2024 The Institute of Electrical and Electronics Engineers, Incorporated} 
}
\usepackage{cite}
\usepackage{amsmath,amssymb,amsfonts}
\usepackage{algorithmic}
\usepackage{graphicx}
\usepackage{textcomp}
\usepackage{xcolor}
\usepackage{booktabs}
\usepackage{comment}
\usepackage{float}
\usepackage[final]{pdfpages}
\usepackage{hyperref}
\usepackage{url}

\usepackage{breakurl}
\usepackage{tablefootnote}
\usepackage{caption}
\usepackage{subcaption}
\usepackage{amsmath}
\usepackage{flushend}
\usepackage{balance}
\usepackage{ulem}
\usepackage{svg}
\usepackage{comment}
\usepackage[bottom]{footmisc}
\usepackage{nameref}
\usepackage[switch]{lineno}

\usepackage{nicematrix}
\usepackage[utf8]{inputenc}
\usepackage{nicematrix}
\usepackage[T1]{fontenc}
\usepackage{mathptmx}
\usepackage{etoolbox}
\usepackage{amsmath,amsfonts}
\usepackage{algorithmic}
\usepackage{textcomp}
\usepackage{booktabs}
\usepackage{comment}
\usepackage{float}
\usepackage[final]{pdfpages}
\usepackage{tablefootnote}
\usepackage{caption}
\usepackage{subcaption}
\usepackage{amsmath}
\usepackage{flushend}
\usepackage{balance}
\usepackage{ulem}
\usepackage[bottom]{footmisc}
\usepackage{tabularx}
\usepackage{colortbl}
\usepackage{xargs}
\usepackage[justification=centering]{caption}

\usepackage{multirow}

\def\BibTeX{{\rm B\kern-.05em{\sc i\kern-.025em b}\kern-.08em
    T\kern-.1667em\lower.7ex\hbox{E}\kern-.125emX}}
\usepackage[switch]{lineno}

\newif\ifdraft
\drafttrue
\ifdraft
  \newcommand{\rajeev}[1]{{\textcolor{red}{ Rajeev: #1 }}}
\else
  \newcommand{\rajeev}[1]{}
\fi
\ifdraft
  \newcommand{\vv}[1]{{\textcolor{orange}{ Venkat: #1 }}}
\else
  \newcommand{\vv}[1]{}
\fi

\definecolor{zhen}{RGB}{0, 0, 0}
\definecolor{zhen_new}{RGB}{0, 143, 67}

\usepackage{multirow}

\begin{document}
\title{LLM-Inference-Bench: Inference Benchmarking of Large Language Models on AI Accelerators}

\makeatletter
\newcommand{\linebreakand}{%
  \end{@IEEEauthorhalign}
  \hfill\mbox{}\par
  \mbox{}\hfill\begin{@IEEEauthorhalign}
}
\makeatother

\author{\IEEEauthorblockN{Krishna Teja Chitty-Venkata\IEEEauthorrefmark{1}\IEEEauthorrefmark{2}\thanks{\IEEEauthorrefmark{2} Equal contribution.}~} \IEEEauthorblockA{schittyvenkata@anl.gov}
\and \IEEEauthorblockN{Siddhisanket Raskar\IEEEauthorrefmark{1}\IEEEauthorrefmark{2}} \IEEEauthorblockA{sraskar@anl.gov}
\and \IEEEauthorblockN{Bharat Kale\IEEEauthorrefmark{1}}  \IEEEauthorblockA{kale@anl.gov}
\and \IEEEauthorblockN{Farah Ferdaus\IEEEauthorrefmark{1}}  \IEEEauthorblockA{fferdaus@anl.gov}
\and \IEEEauthorblockN{Aditya Tanikanti\IEEEauthorrefmark{1}}  \IEEEauthorblockA{atanikanti@anl.gov} 
\and \IEEEauthorblockN{Ken Raffenetti\IEEEauthorrefmark{1}} \IEEEauthorblockA{raffenet@anl.gov}
\and \IEEEauthorblockN{Valerie Taylor\IEEEauthorrefmark{1}} \IEEEauthorblockA{vtaylor@anl.gov}
\and \IEEEauthorblockN{Murali Emani\IEEEauthorrefmark{1}} \IEEEauthorblockA{memani@anl.gov}
\and  \IEEEauthorblockN{Venkatram Vishwanath\IEEEauthorrefmark{1}} \IEEEauthorblockA{venkat@anl.gov}
  
  \linebreakand
  
\IEEEauthorrefmark{1}Argonne National Laboratory, Lemont, IL 60439, USA 
}
\maketitle


\raggedbottom

\begin{abstract}

Large Language Models (LLMs) have propelled groundbreaking advancements across several domains and are commonly used for text generation applications. However, the computational demands of these complex models pose significant challenges, requiring efficient hardware acceleration. Benchmarking the performance of LLMs across diverse hardware platforms is crucial to understanding their scalability and throughput characteristics. We introduce LLM-Inference-Bench, a comprehensive benchmarking suite to evaluate the hardware inference performance of LLMs. We thoroughly analyze diverse hardware platforms, including GPUs from Nvidia and AMD and specialized AI accelerators, Intel Habana and SambaNova. Our evaluation includes several LLM inference frameworks and models from LLaMA, Mistral, and Qwen families with 7B and 70B parameters. Our benchmarking results reveal the strengths and limitations of various models, hardware platforms, and inference frameworks. We provide an interactive dashboard to help identify configurations for optimal performance for a given hardware platform. 

\end{abstract}



\pagestyle{plain}
\thispagestyle{plain}

\section{Introduction}
\label{sec:intro}

\textit{Large Language Models} (LLMs) have emerged as a transformative force in Artificial Intelligence (AI), revolutionizing Natural Language Processing (NLP) and text generation. These models, such as GPT \cite{brown2020language}, LLaMA \cite{dubey2024llama3herdmodels}, and LaMDA \cite{thoppilan2022lamda}, have risen to prominence due to their ability to understand and generate human-like text across various tasks. LLMs are now being utilized in various applications, including content generation, question-answering, and language translation in several domains including scientific machine learning. However, the deployment and usage of LLMs come with significant computational challenges. Training these massive models requires substantial computational resources, leading to increased energy consumption and environmental concerns. As the model complexity grows by leaps and bounds, several innovative solutions have been developed to address these challenges, including improved AI hardware infrastructure, cost-efficient training techniques, memory-efficient fine-tuning methods, and optimized inference solutions.  

\textit{LLM Inference} \cite{chitty2023survey, zhou2024survey, park2024inference} is a critical aspect of various modern applications, which refers to using a trained LLM to generate responses or make predictions. Today, efficient inference is essential for generation capabilities across various applications, such as chatbots, language translation, and information retrieval systems. As LLMs continue to grow in size and complexity, optimizing inference becomes increasingly crucial to balance performance with computational resources, energy consumption, and response times. 

In recent years, the development of hardware accelerators for Deep Learning (DL) applications, such as GPUs and TPUs, has been driven to meet the computational demands of large models. These accelerators are designed to enhance performance and energy efficiency, which is particularly crucial for LLMs that consist of billions of parameters. These hardware solutions significantly improve performance, including faster training times, reduced inference latency, and enhanced scalability. This is essential for developing and deploying sophisticated models capable of handling state-of-the-art (SOTA) tasks in NLP, content generation, and decision support systems. The advancements in AI hardware platforms directly impact the ability to develop more complex models, especially for LLMs. 

The landscape of LLMs has evolved significantly in recent years, marked by three parallel trends: the proliferation of open-source LLMs, development of specialized AI accelerators and efficient inference frameworks. These models, along with efficient accelerators and optimized software, aim to enhance LLM performance, making hardware inference benchmarking crucial for identifying bottlenecks and maximizing efficiency. Benchmarks provide a standardized methodology for comparing various hardware platforms, model architectures, and inference optimizations, revealing trade-offs between factors such as latency, throughput, and energy consumption. This allows users to optimize based on their priorities and gain a comprehensive understanding of the capabilities and limitations of different AI accelerators. In this paper, we introduce LLM-Inference-Bench, a comprehensive benchmarking suite designed to provide detailed performance evaluations of LLMs across multiple AI accelerators, contributing to the broader understanding of LLM performance optimization and hardware selection in the rapidly evolving field of AI acceleration. This comparative study can serve as a resource for researchers seeking to optimize LLMs based on their specific hardware, framework and model requirements.

The main contributions of our paper are as follows: 

\begin{enumerate}
    \item We introduce \textit{LLM-Inference-Bench}, a comprehensive benchmarking study that evaluates the inference performance of the LLaMA model family, including LLaMA-2-7B, LLaMA-2-70B, LLaMA-3-8B, LLaMA-3-70B, as well as other prominent LLaMA derivatives such as Mistral-7B, Mixtral-8x7B, Qwen-2-7B, and Qwen-2-72B across a variety of AI accelerators, including Nvidia A100, Nvidia H100, Nvidia GH200, AMD MI300X and AMD MI250 GPU, as well as specialized AI accelerators SambaNova SN40L and Habana Gaudi2. 
    
    \item We provide comprehensive data on hardware performance metrics such as \textit{throughput} and \textit{power consumption} for different inference frameworks like vLLM, TensorRT-LLM, llama.cpp, and Deepspeed-MII across systems, where supported. In addition, we report validation metrics such as perplexity on selected benchmarks. For a broader community reach, we also open source interactive LLM-Inference-Bench Dashboard along with artifacts to reproduce all results. The code and dashboard for this paper can be accessed here: \href{https://github.com/argonne-lcf/LLM-Inference-Bench}{https://github.com/argonne-lcf/LLM-Inference-Bench} 
    
    \item Our study examines multiple factors that significantly influence the inference performance of an LLM, which include the model parameter size, input sequence length, number of output tokens generated and batch size. Based on our results, we provide key insights and takeaways from three different perspectives: framework-wise, accelerator-wise and model-wise. 
    

\end{enumerate}

\section{Background and Related Work}
\label{sec:background}

This section presents a concise overview of the fundamental concepts of LLMs. This background information is essential to understand key performance differences among models and will be helpful for the subsequent discussions on hardware and framework comparison.

\subsection{LLM Architecture} 
LLMs are primarily based on the traditional transformer architecture \cite{vaswani2017attention}, consisting of multiple encoder layers that can analyze input text data and decoder layers for generating output. Their core components include input embeddings, positional encoding, self-attention, and feed-forward layers. This paper considers only decoder-based LLMs, which generate new tokens as output based on the input prompt.  

\textbf{Dense LLMs} represent the conventional approach, utilizing a single, extensive neural network architecture. In these models, all parameters are actively engaged during each inference or training pass, with network operations occurring sequentially and interconnectedly. This architecture allows for comprehensive information processing but can be computationally intensive. Notable examples include LLaMA-2-7B \cite{touvron2023llama}.

\textbf{Mixture of Experts} (MoE) LLMs \cite{shazeer2017outrageously, cai2024survey} employ multiple specialized sub-networks with a routing mechanism, activating only relevant experts for each input. This approach offers improved parameter efficiency and potentially faster inference than dense models. MoE architectures enable larger, more scalable models with specialized knowledge but introduce additional complexity in training and deployment. In MoE models, such as Mixtral-8x7B \cite{mixtral_8_7b}, the usage of different experts is within the MLP block, as depicted in Figure \ref{fig:Mixtral_MoE}. 


\textbf{MHSA vs GQA} There exists different types of attention operations and the most prominent of them are Multi-Head Self-Attention (MHSA) and Group Query Attention (GQA). In an MHSA module, each attention head consists of unique query, key, and value vectors, as depicted in Figure \ref{fig:mhsa}. This allows the model to attend to different subspaces of the input representation in parallel. MHSA offers the best performance but is computationally expensive and memory-intensive for large models. GQA divides query heads into multiple groups where each group shares a single key head and value head, as depicted in Figure \ref{fig:gqa}. GQA has \#groups times fewer parameters than MHSA due to KV head sharing.

\subsection{Benchmarking}

Evaluation of LLMs on diverse hardware platforms is of crucial importance to understand the capabilities and limitations of both traditional and non-traditional architectures. Prior work has studied LLMs on leadership class supercomputers \cite{emani2023gpt2, llmonfrontier} and with traditional deep learning benchmarks \cite{emani2022cnn, YIN2021100005} providing detailed insights into their capabilities. To the best of our knowledge, our work is the first paper to provide dedicated inference benchmarking results and analysis targeting a wide range of SOTA hardware, models and frameworks.

\section{Experimental Setup}
\label{sec:setup}

Our study aims to benchmark LLMs across various AI accelerators and inference frameworks comprehensively. This section details the LLM architectures, hardware configurations, and inference frameworks used in our experiments.

\subsubsection{\textbf{LLM Architectures}} 
We choose a diverse set of SOTA LLMs for benchmarking, ranging from 7 to 70 billion parameters. The models included in the paper include LLaMA-2-7B \cite{touvron2023llama}, LLaMA-2-70B \cite{touvron2023llama}, LLaMA-3-8B \cite{dubey2024llama3herdmodels}, LLaMA-3-70B \cite{dubey2024llama3herdmodels}, Mistral-7B \cite{jiang2023mistral}, Mixtral-8x7B \cite{jiang2024mixtral}, Qwen-2-7B \cite{yang2024qwen2} and Qwen-2-72B \cite{yang2024qwen2}. These models represent a mix of architectures, including traditional decoder-only and MoE models. We aim to provide insights into how different model architectures affect inference performance across hardware platforms using these diverse models. Please refer to Table \ref{table:model_Summary} (Appendix \ref{appendix:llms}), where we summarize the neural architecture configurations of different LLMs.

\subsubsection{\textbf{LLM Token Generation Parameters}} 
\textit{Input Length} is the number of tokens given to an LLM as input prompt for a single query.  
\textit{Output Size}, also referred to as \texttt{max\_new\_tokens} (maximum new tokens) or output generated tokens, is the number of tokens produced by the model as a response to the input prompt. The output generation is an iterative process where the model predicts and appends the token one at a time until it reaches a stopping condition or a specified token limit. \textit{Batch Size} refers to the number of input sequences processed and new output sequences produced simultaneously. In this work, we consider input and output lengths of 128, 256, 512, 1024, and 2048 and batch sizes of 1, 16, 32, and 64.

\subsubsection{\textbf{Hardware Configurations}} 

We deploy LLMs on various hardware to benchmark their inference performance. We choose various SOTA GPUs from vendors like Nvidia (A100 SXM-40GB \cite{nvidia_a100}, H100 SXM5 80GB\cite{choquette2023nvidia}, GH200 \cite{nvidia_gh200}) and AMD MI250 \cite{amd_mi250}, MI300X \cite{amd_mi300x}. We extend our study on specialized AI accelerators like Intel Habana Gaudi2 \cite{habana2022gaudi2} and Sambanova SN40L \cite{prabhakar2024sambanova}. Please refer to Table \ref{table:hwoverview} (Appendix \ref{appendix:platforms}), where we summarize the hardware features of these evaluated systems. 

\subsubsection{\textbf{Inference Frameworks}} 
We evaluate the inference performance of LLMs on the aforementioned hardware on the following SOTA inference frameworks: \textbf{TensorRT-LLM} (TRT-LLM) \cite{trtllm_repo} is Nvidia's inference library optimized for LLMs which provides high throughput and low latency. It is designed and optimized for NVIDIA GPUs by leveraging TensorRT, CUDA and cuDNN libraries to accelerate LLM inference. Therefore, TensorRT-LLM can be used only to accelerate LLMs on NVIDIA GPUs. We used the TensorRT-LLM pip version of 0.11.0 in our experiments. \textbf{vLLM} \cite{kwon2023efficient} is an open-source and community-maintained library known for its efficient memory management and support across a wide range of accelerators. \textbf{Deepspeed-MII} (DS-MII) \cite{ds-mii} is Microsoft's model implementation for LLM Inference, built on the DeepSpeed library \cite{aminabadi2022deepspeed} known for large-scale inference. \textbf{llama.cpp} \cite{llama_cpp} is a lightweight framework for running LLMs, written in C/C++, and is known for its efficiency and portability across various hardware/software configurations, including CUDA, OpenCL, and Metal. Please refer to Appendix \ref{appendix:infr_frameworks} for a detailed description of each inference framework.


\subsubsection{\textbf{Performance Metrics}} 

We consider the following validation and performance metrics in our paper:

\textbf{(a) Perplexity} quantifies the model's level of surprise when encountering new data to generate a new token. A lower perplexity indicates better performance and is calculated as an exponent of the model's loss, measuring how well an LLM has learned to generate new text. 

\textbf{(b) Time to First Token (TTFT)} is the amount of time required to produce the first output token after receiving an input prompt. It represents how quickly the users can see the LLM's output after submitting their query. We measure TTFT by setting the maximum output to one token and recording the time to generate this output. 

\textbf{(c) Inter Token Latency (ITL)} refers to the average time interval between generating consecutive tokens. It measures how quickly the model can produce each subsequent token after the previous one.
\begin{equation} \label{eq:ITL}
\text{ITL} = \frac{(\text{End-to-End Latency} - \text{TTFT})}{\text{Batch Size x (Output Tokens-1)}}
\end{equation}

\textbf{(d) Throughput} is a key indicator of a hardware's processing efficiency. It provides insight into the model's capacity to handle sequences and batches. In this paper, we define throughput as the total number of tokens (both input and output) processed by the hardware per second. We first calculate the end-to-end latency, the time elapsed between the input prompt provided to LLM, and the generation of the final output token. We convert latency to throughput using Equation \ref{eq:throughput}.
\begin{equation} \label{eq:throughput}
\text{throughput} = \frac{\text{Batch Size} \times (\text{Input} + \text{Output Tokens})}{\text{End-to-End Latency}}
\end{equation}

\textbf{(e) Power:} LLM inference requires substantial computational resources, leading to increased energy consumption, making power benchmarking quintessential. We report the power consumed only by the accelerators, not host and other peripherals. We use \textit{Average Power} as a performance metric, which is calculated as the ratio of total work done to the total time taken and is measured in watts. Also, we compare the \textit{performance per watt} (measured in tokens/sec/watt) across GPUs, which is the number of tokens processed per second for unit consumption of power. This paper reports power metrics of only Nvidia GPUs using \texttt{pynvml} \cite{pynvml} and these measurements on other hardware are planned for future work.

\section{LLM Inference - Preliminary Study}
\label{sec:infr_prelim_study}

The deployment of LLMs into production environments demands efficient inference capabilities. This section first examines the role of input/output sequence lengths and batch sizes, followed by delving into different algorithm and hardware optimization strategies to understand different approaches.

\subsection{Varying Input, Output, and Batch Sizes}

\subsubsection{\textbf{Dynamic Batch Sizes}} In general, LLMs demonstrate increasing throughput with an increase in batch sizes for the same input and output length until the compute and memory resources of the parallel hardware are fully saturated. This is due to the simultaneous execution of all input sequences and parallel output token generation of batches. Frameworks like vLLM, TensorRT-LLM and accelerators such as H100, SN40L use continuous batching \cite{yu2022orca}, a dynamic batching strategy to process multiple requests concurrently, even if the requests arrive at different times or have different input context lengths. This method keeps the device busy, and new requests of variable length can be processed without waiting for the previous batch to be finished.

\begin{figure}[H]
    \centering
    \begin{subfigure}[b]{0.48\linewidth}
        \centering
        \includegraphics[width=1.05\linewidth]{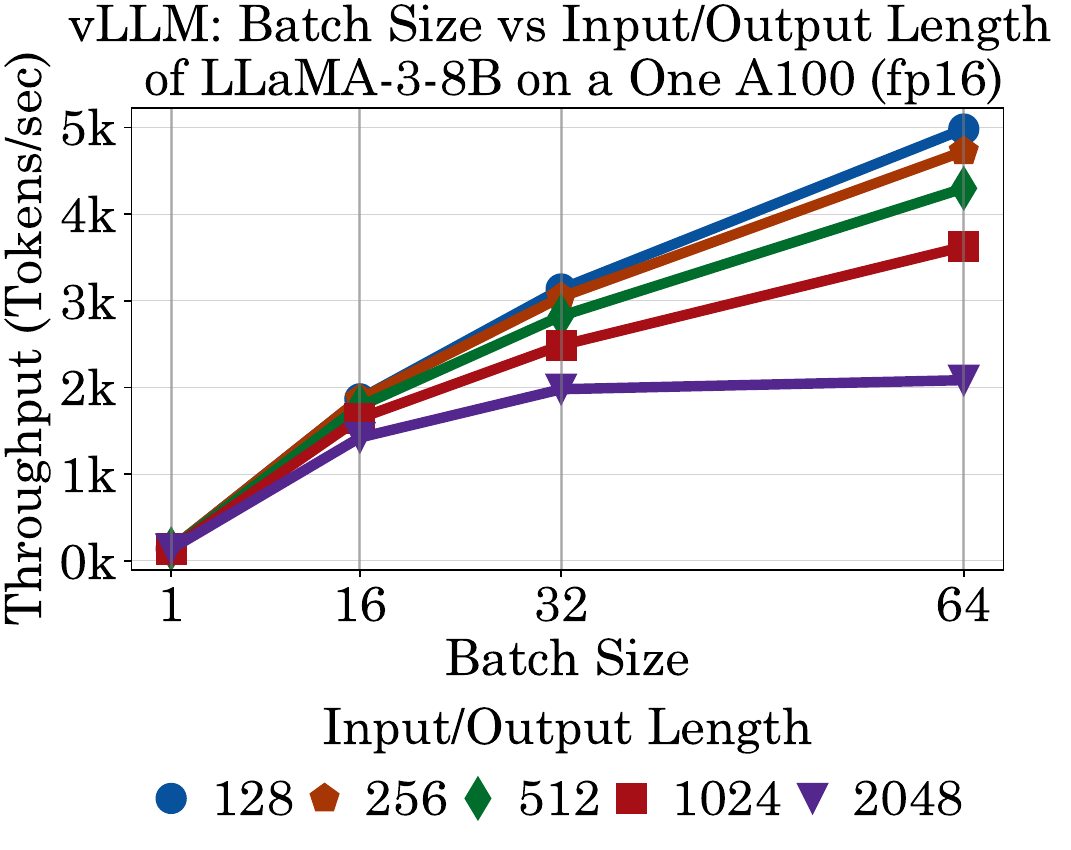}
        \caption{vLLM: Batch Size vs Input Output Length}
        \label{fig:Batch_Size_vs_Seq_Length}
    \end{subfigure}
    \begin{subfigure}[b]{0.48\linewidth}
        \centering
        \includegraphics[width=\linewidth]{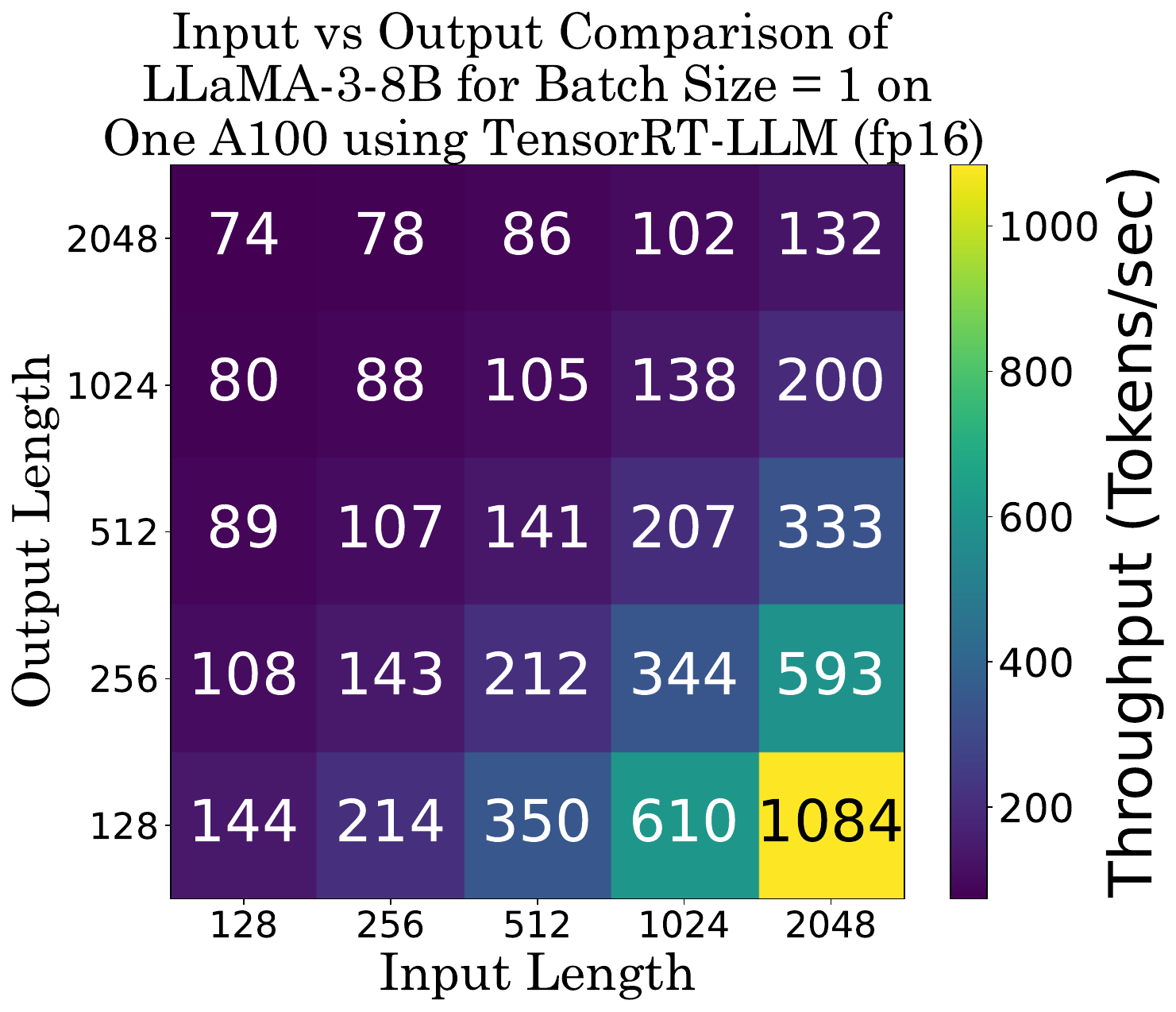}
        \caption{TRT-LLM: Input vs Output Length}
        \label{fig:inp_out_heatmap}
    \end{subfigure}
    \caption{LLaMA-3-8B on single A100
    }
    \label{fig:batch_size_inpu_length}
\end{figure}

Figure \ref{fig:Batch_Size_vs_Seq_Length} shows that the throughput increases with an increase in batch size for the input/outut length. For a batch size of 64, the throughput is 26.6x greater than that of a batch size of 1 for a token length of 2048 on A100. We see increasing throughput with increasing batch sizes due to GPU processing more input sequences simultaneously and fixed costs such as kernel launch times occurring only once for the entire batch.

\subsubsection{\textbf{Input/Output Sizes}} Blended tokens are defined as a situation where the input size differs from the output tokens, such as summarization and text classification, which require outputs significantly smaller than the input token length and text completion and code generation, which require outputs longer than the input prompt. For a given batch size, throughput increases as the output length decreases for the same input size. In contrast, it decreases as the input length decreases for the same output length due to the sequential nature of output token generation compared to the parallel processing of input tokens. The heatmap in Figure \ref{fig:inp_out_heatmap} shows that the throughput for an \{input, output\} size of \{1024, 128\} is 14.6 times greater than for an \{input, output\} size of \{128, 1024\}.

\raggedbottom

\subsection{Algorithm Optimizations}

\subsubsection{\textbf{KV Cache}}

LLMs generate text \textit{autoregressively}, i.e., they produce only one token as output based on all the previous tokens. The new token generated is appended to the input to produce future tokens. KV Caching \cite{pope2023efficiently, luohe2024keep} improves the efficiency of LLM inference by reusing the past Key-Value pairs, thereby eliminating the need for recalculation for every new token. Without using this KV cache, the model must recompute attention heads for all previous tokens for new token generation. Figure \ref{fig:w_w_o_KV_Cache} depicts the performance of 70B models with and without KV caching on Gaudi2 (8 HPUs). The results indicate a substantial improvement ($\sim$2x for 128 and $\sim$7x for 1024 length) in throughput with KV caching.

\begin{figure}[H]
    \centering
    \begin{subfigure}[b]{0.48\linewidth}
        \centering
        \includegraphics[width=\linewidth]{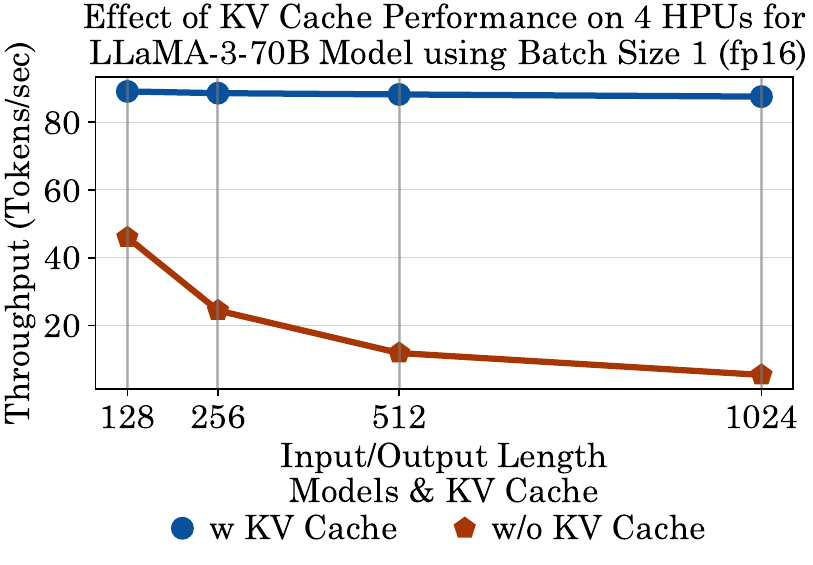}
        \caption{with \& without KV Cache on Gaudi2 with Batch Size 1}
        \label{fig:w_w_o_KV_Cache}
    \end{subfigure}
    \begin{subfigure}[b]{0.48\linewidth}
        \centering
        \includegraphics[width=\linewidth]{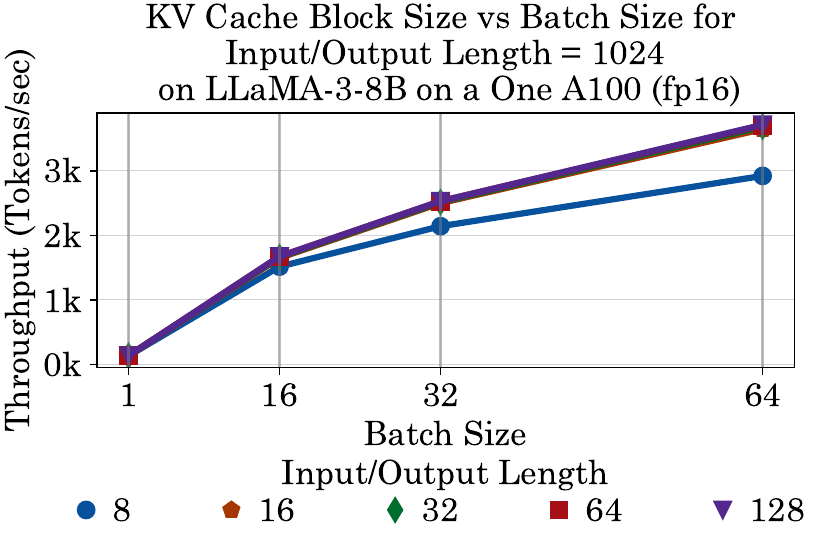}
        \caption{KV Cache Block Size vs Batch Size}
        \label{fig:KV_Cache_block_size}
    \end{subfigure}
    \caption{KV Cache Performance Benchmarking}
    \label{fig:KV_Cache_Performance_Benchmarking}
\end{figure}

\subsubsection{\textbf{Blocked KV Caching}} vLLM \cite{kwon2023efficient} addresses the challenge of memory fragmentation in LLM serving to facilitate non-contiguous caching. Traditional KV caches in LLMs are monolithic and variable-sized, leading to memory fragmentation and reduced concurrency. Inspired by OS virtual memory, vLLM uses fixed-sized blocks or pages instead of variable-sized contiguous chunks. This blocking increases throughput by eliminating memory fragmentation. Figure \ref{fig:KV_Cache_block_size} depicts that on an A100 GPU, any KV cache block size greater than or equal to 16 produces optimal throughput, while low block sizes hurt the performance. For a batch size of 64, the throughput for block size 16 is 1.27x greater than block size 8.

\subsubsection{\textbf{Quantization}} Quantization \cite{gholami2022survey} is a popular method for model size reduction by lowering the precision of weight and activations. LLM Quantization is extremely critical, given the sheer size and complexity. LLMs can be operated in lower precisions such as FP8 \cite{kuzmin2022fp8}, using GPTQ \cite{frantar2022gptq} and AWQ \cite{lin2023awq} without compromising the output quality. Figure \ref{fig:llama_3_quant} compares FP16, FP8 and Int8 precision using vLLM and TRT-LLM on A100 and H100. We observe that FP8 on H100 and Int8 on A100 can provide performance benefit compared to FP16 and the absence of FP8 support on A100 limits the framework's ability to leverage low precision for weights and KV cache.

\begin{figure}[H]
        \centering
        \includegraphics[width=1\linewidth]{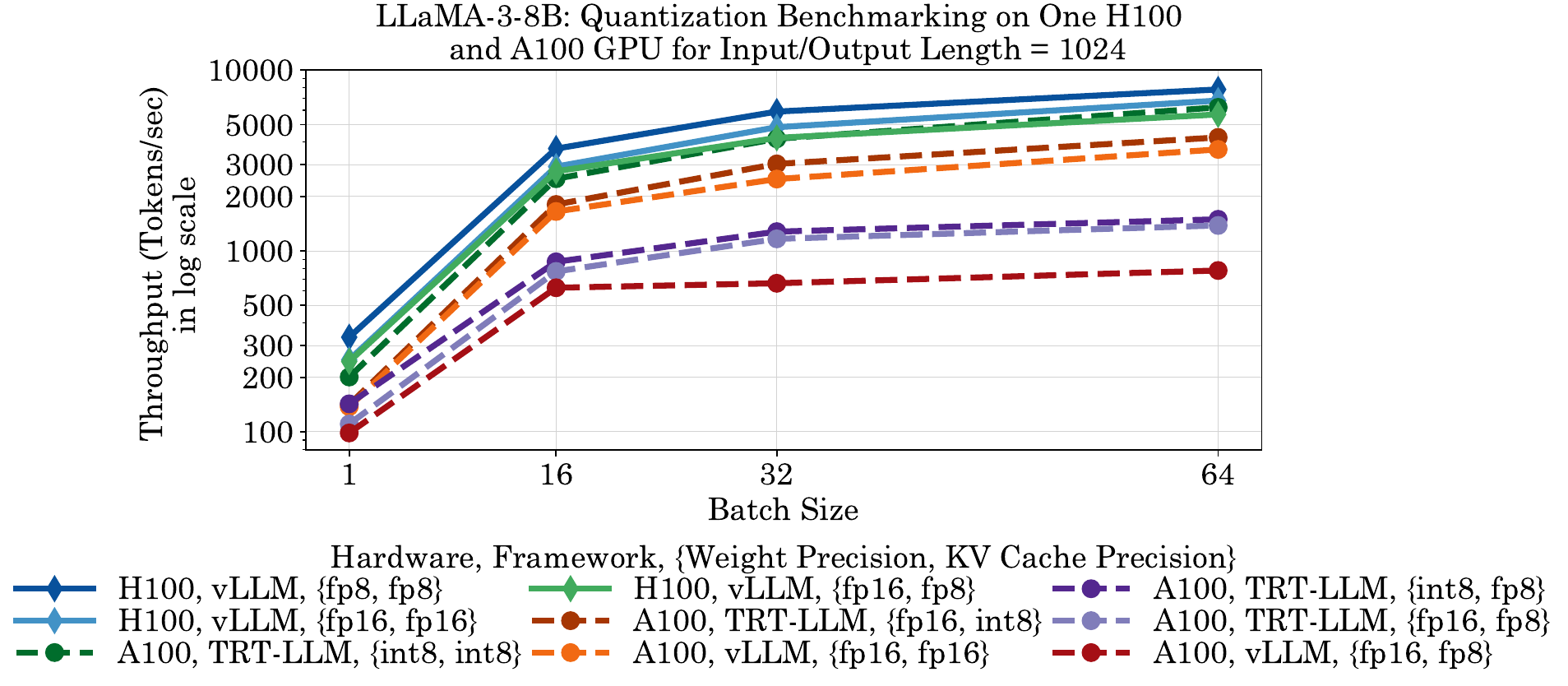}
        \caption{LLaMA-3-8B Quantization Benchmarking}
        \label{fig:llama_3_quant}
        \captionsetup{justification=centering}
\end{figure}

\subsubsection{\textbf{Neural Architecture Search (NAS)}} NAS \cite{chitty2022neural, chitty2022neural_transformers} is an automated process for discovering optimal neural architectures tailored to specific datasets and hardware constraints. It can be applied extended to LLMs to optimize the architectural elements. DeciLM-7B \cite{decilm_7b} utilizes NAS to determine the optimal KV head sizes in each layer from the pool of the following options: \{1,2,4\}. The searched Deci model has 67 KV heads across all 32 layers, while LLaMA-3-8B and Mistral-7B have 256 (8*32) KV heads throughout the model. Figure \ref{fig:NAS} shows the performance benefit of DeciLM-7B over LLaMA-3-8B and Mistral-7B on A100 and H100 GPUs. 

\begin{figure}[H]
    \centering
    \begin{subfigure}[b]{0.48\linewidth}
        \centering
        \includegraphics[width=\linewidth]{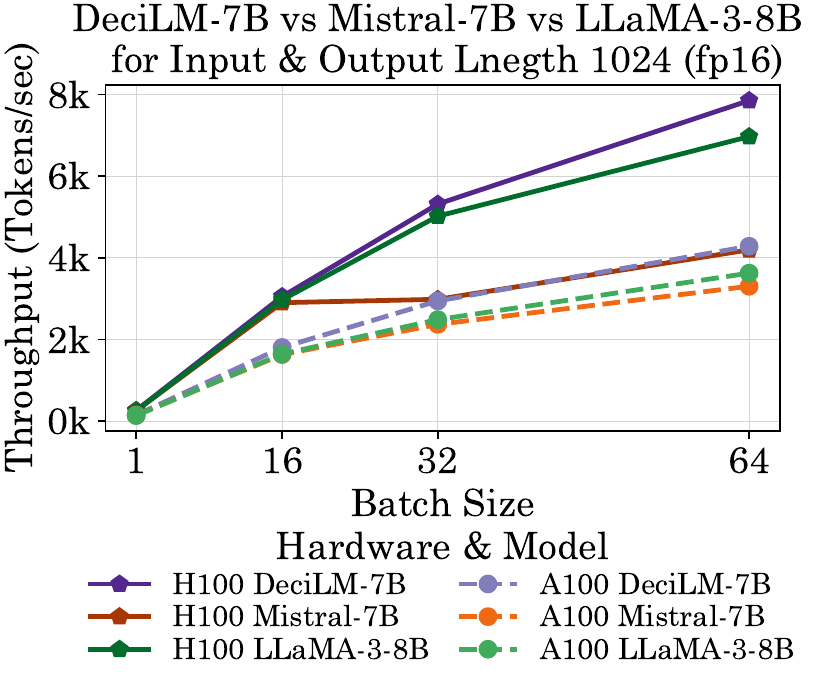}
        \caption{NAS vs non-NAS Models}
        \label{fig:NAS}
    \end{subfigure}
    \begin{subfigure}[b]{0.48\linewidth}
        \centering
        \includegraphics[width=\linewidth]{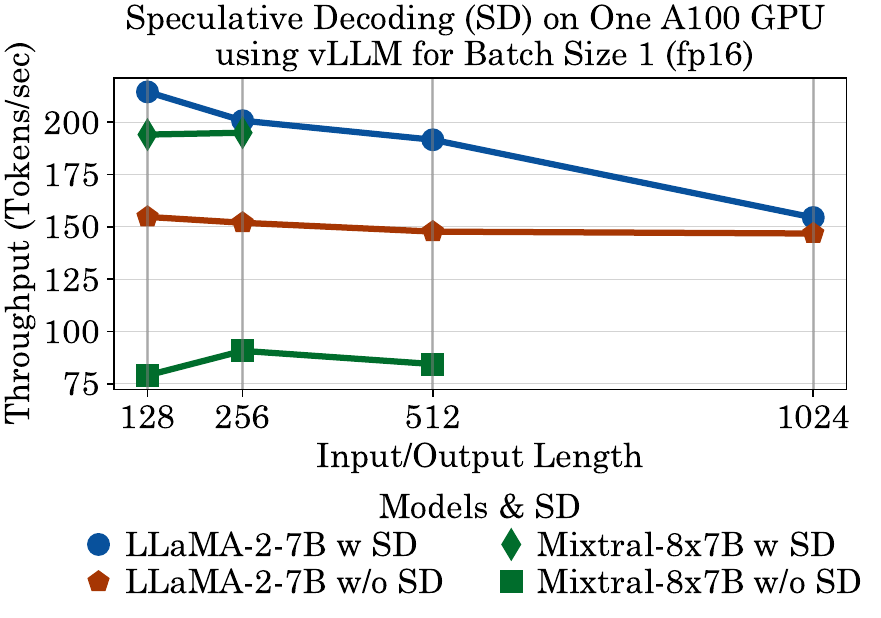}
        \caption{SD using vLLM}
        \label{fig:SD}
    \end{subfigure}
    \caption{NAS and SD on A100 GPU}
    \label{fig:NAS_SD}
\end{figure}

\subsubsection{\textbf{Speculative Decoding (SD)}} SD \cite{xia2024unlocking} is a technique which involves a dual-model system comprising a target model, a larger LLM intended for the main task, and a small draft model, a lightweight LLM. The small draft model generates initial token guesses, verified and corrected by the main model. However, with an increase in sequence length and model size, the benefit of SD vanishes, as depicted in Figure \ref{fig:SD}. We used the LLaMA-68M model \cite{llama_68m} as a draft network for LLaMA-2-7B and Mixtral-8x7B and observed that SD improves the performance of only the 7B model.

\subsection{Parallelism Techniques}

\subsubsection{\textbf{Tensor Parallelism} (TP)} TP \cite{shoeybi2019megatron} is a technique to distribute the weight tensor of a layer across multiple devices, either row-wise or column-wise. The devices communicate with each other to share the input and output activations. Tensor parallel is very effective within a single node as inter-node data transfer is quicker than intra-node communication. This allows distributing the memory and computation for large tensors that wouldn't fit on a single device.

\subsubsection{\textbf{Pipeline Parallelism} (PP)} In PP \cite{huang2019gpipe}, the model is divided into different layers, and each device computes its assigned layers and passes the output to the next device in the pipeline. The layer's output must be transferred across GPUs, whereas weights and KV cache can be local to the device.

\subsubsection{\textbf{Expert  Parallelism} (EP)} EP \cite{rajbhandari2022deepspeed} distributes MoE models across multiple devices. This method leverages the independent nature of experts in the MoE layers and assigns a group of expert blocks on a single device. A load balancing issue may exist when experts assigned to a GPU are not active. 

\subsubsection{\textbf{Hybrid Parallelism} (HP)} HP \cite{singh2023hybrid} combines multiple parallelisms, such as TP, PP and EP, to allow efficient scaling and better hardware utilization. This method offers greater flexibility as different parallelism techniques can be applied to the layer. However, managing multiple parallelisms simultaneously can be complex as the distribution of work across all devices becomes more challenging.

\begin{figure}[H]
    \centering
    \begin{subfigure}[b]{0.48\linewidth}
        \centering
        \includegraphics[width=\linewidth]{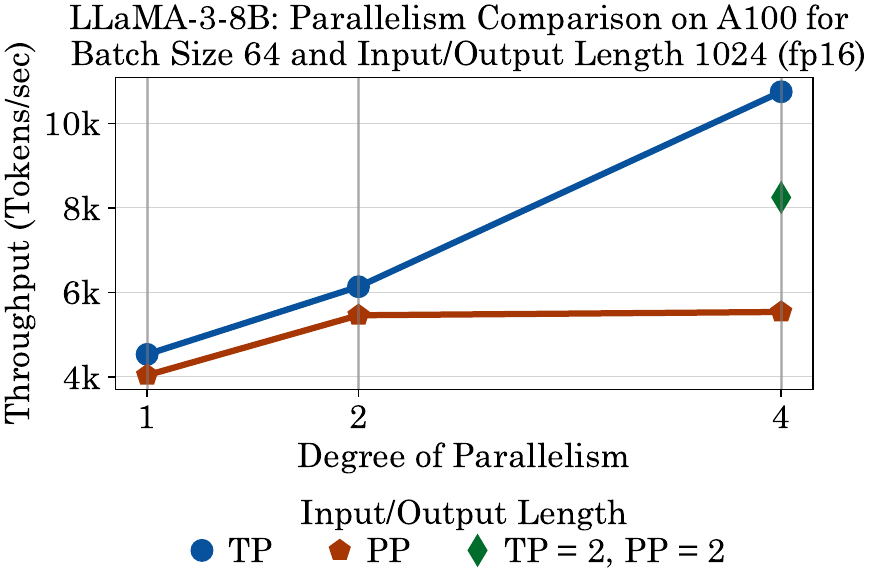}
        \caption{TP and PP on LLaMA-3-8B}
        \label{fig:tp_pp_llama_3_8b}
    \end{subfigure}
    \begin{subfigure}[b]{0.48\linewidth}
        \centering
        \includegraphics[width=\linewidth]{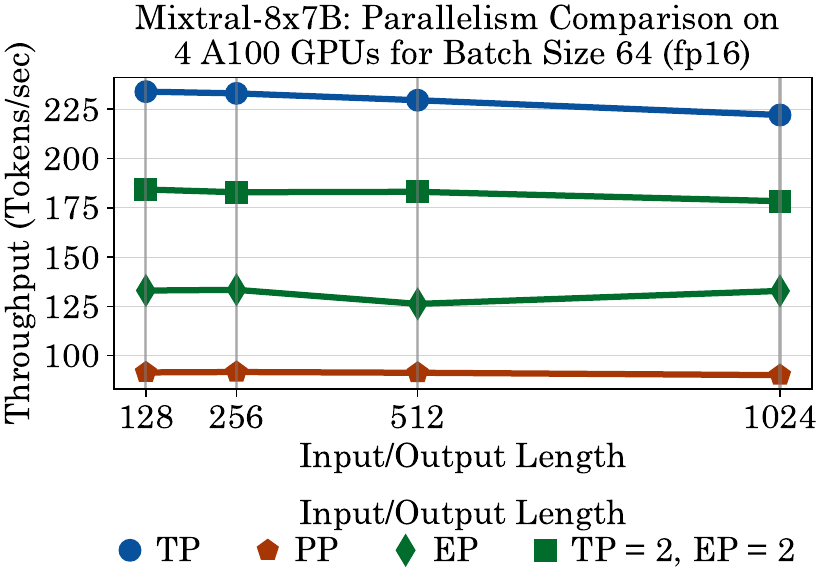}
        \caption{TP, PP, EP on Mixtral-7x8B}
        \label{fig:tp_pp_ep_mixtral}
    \end{subfigure}
    \caption{Parallelism Comparison within a node}
    \label{fig:TP_PP_EP_Hybrid}
\end{figure}

Figure \ref{fig:tp_pp_llama_3_8b} depicts the performance of LLaMA-3-8B on 1,2 and 4 GPUs, and Figure \ref{fig:tp_pp_ep_mixtral} illustrates the comparison of Mixtral-8x7B using different parallelism within a single node. While the hybrid parallelism approach offers great flexibility, TP is effective due to more device utilization and less communication overhead. Our observations indicate that. TP is 1.30x faster than the hybrid approach (TP=2,PP=2) and 1.94x faster than PP on 4 A100 GPUs using LLaMA-3-8B.

\section{Framework-wise Benchmarking}

In this section, we perform framework-wise benchmarking, which helps to compare different models and accelerators. We focus on unique features, capabilities, and performance characteristics. In all the results below, we used 16 bits, and the number of GPUs is equal to the TP size.  


\subsubsection{\textbf{TensorRT-LLM}} TRT-LLM execution involves three steps: 1) Converting pretrained HuggingFace model to TensorRT-LLM checkpoints, considering tensor parallelism and batch size, 2) Building an optimized binary, and 3) Executing the binary on NVIDIA GPUs. In Figure \ref{fig:trt_7b_models_across_batch_size}, we compare the performance of $~$7B models on one GH200, H100 and A100 GPU. The results show that the newer generation of Nvidia GPU outperforms the previous generation. The performance improvement of LLaMA-2-7B plateaus compared to Mistral-7B and LLaMA-3-8B for large batch sizes. This is mainly due to the GQA implementation in the latter two architectures. GQA requires less computation and KV cache memory, and this operation is optimized well in this framework. GQA models (Mistral-7B and LLaMA-3-8B) are approximately 1.9x and 2.79x faster than LLaMA-2-7B on H100 and A100, respectively, for batch size 64. The minimal performance difference between Mistral-7B and LLaMA-3-8B is due a larger vocab size in the latter model.

\begin{figure}[H]
    \centering
     \includegraphics[width=\linewidth]{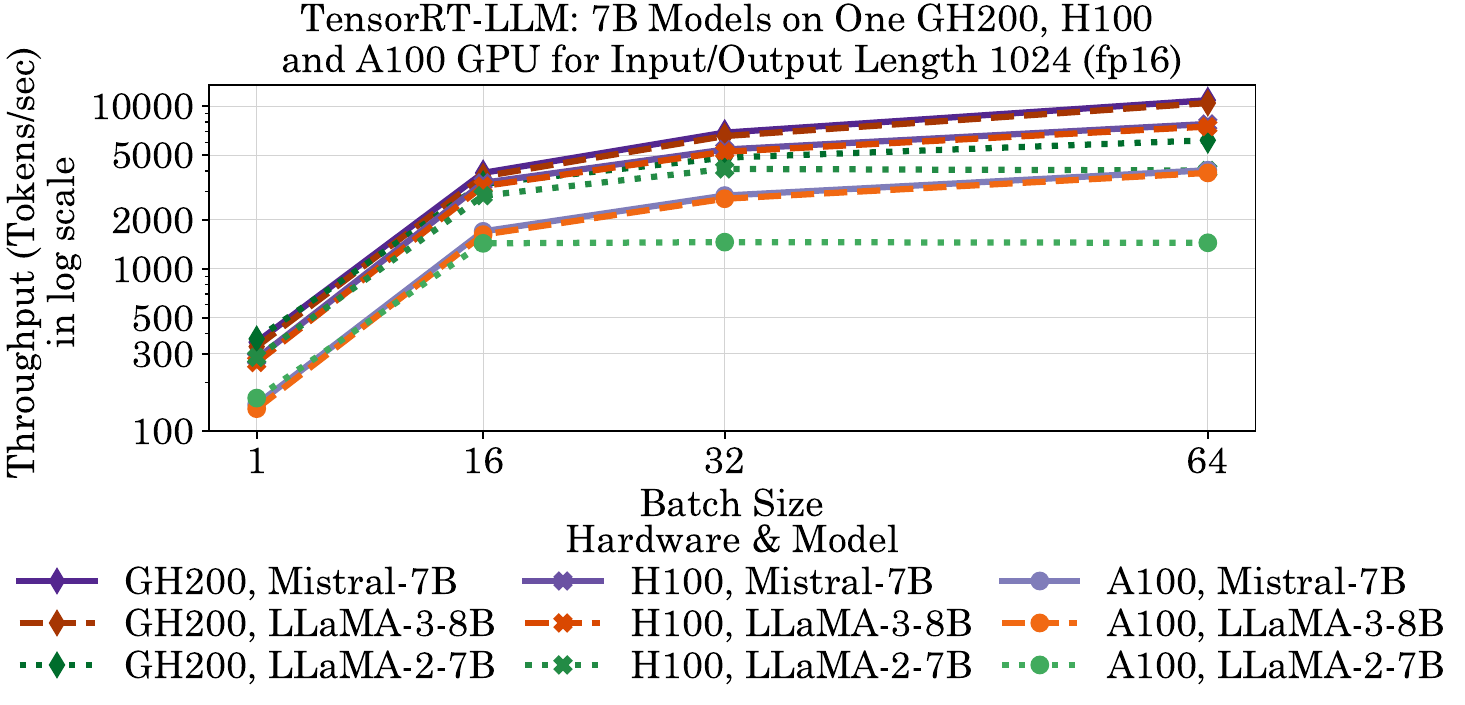}
        \caption{Throughput of 7B Models using TRT-LLM}
        \label{fig:trt_7b_models_across_batch_size}
    \captionsetup{justification=centering}
\end{figure}

\begin{figure}[H]
    \centering
    \includegraphics[width=\linewidth]{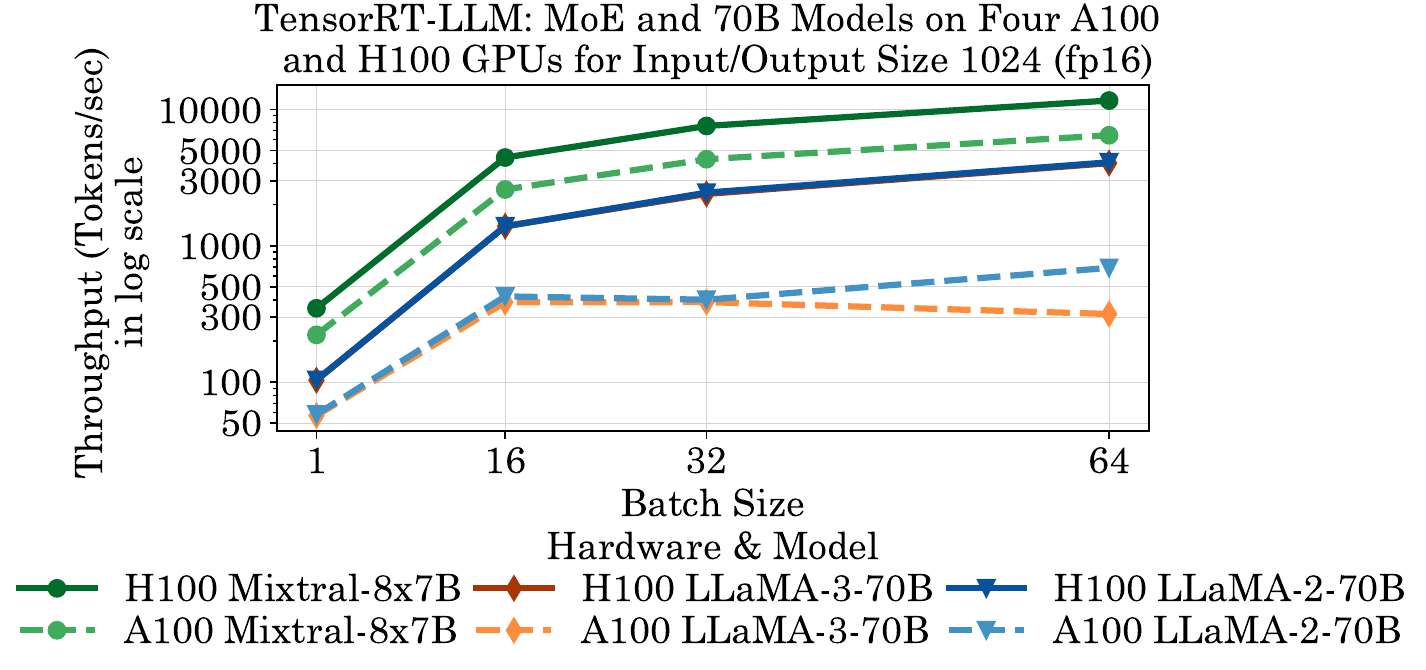}
     \captionsetup{justification=centering}
     \caption{Throughput of 70B/MoE Models using TRT-LLM}
     \label{fig:trt_70b_models_across_batch_size}
\end{figure}

Figure \ref{fig:trt_70b_models_across_batch_size} compares the performance of Mixtral-7x8B and 70B models on H100 and A100 GPUs. The Mixtral model outperforms 70B models, whereas LLaMA-2-70B outperforms LLaMA-3-70B due to less vocabulary size. The Mixtral model is equivalent to a 14B model, as only two of eight experts are active per layer during inference. The H100 GPU demonstrates a more significant performance boost over the A100, especially at larger batch sizes, due to the improved transformer engine \cite{nvidia_h100}, tensor core architecture, and higher memory bandwidth. For a batch size of 64, LLaMA-3-70B on H100 is $~$7.8x faster than A100. Also, H100 GPU scales efficiently with an increase in batch size due to large DRAM and tensor core utilization compared to A100. The throughput of LLaMA-3-70B on H100 improves by a factor of 39x when increasing the batch size from 1 to 64 as opposed to 3x on A100.

\subsubsection{\textbf{vLLM}} vLLM is highly flexible and can be accelerated over various hardware platforms compared to TensorRT-LLM. vLLM better utilizes CUDA kernels for efficient memory management to manage KV memory (PagedAttention). Figure \ref{fig:vLLM_7B_across_hardware} shows that vLLM on GH200 consistently achieves the highest throughput across all batch sizes, and H100 is the second-best closest performer. This is due to 3.5x more memory and tight coupling of Grace CPU and Hopper GPU on a single package on GH200 GPU, providing more bandwidth than H100 GPU. A100 and MI250 show similar performance across models, with A100 marginally ahead. Qwen2-7B on GH200 has the highest throughput compared to other $~$7B models on other hardware. This is due to a smaller hidden size, attention heads and \#layers in Qwen-2-7B than in other 7B models. LLaMA-3-8B exhibits higher throughput than LLaMA-2-7B on the same GPU for large batch sizes despite one billion more parameters due to GQA in the former model. Figure \ref{fig:vLLM_70B_across_hardware} compares the performance difference between 70B models. The trend follows similar to TRT-LLM, where LLaMA-2-70B is faster than LLaMA-3-70B and Qwen-2-72B. Also, the Mixtral-8x7B model performs better than the 70B models.

\begin{figure}
     \centering
        \includegraphics[width=\linewidth]{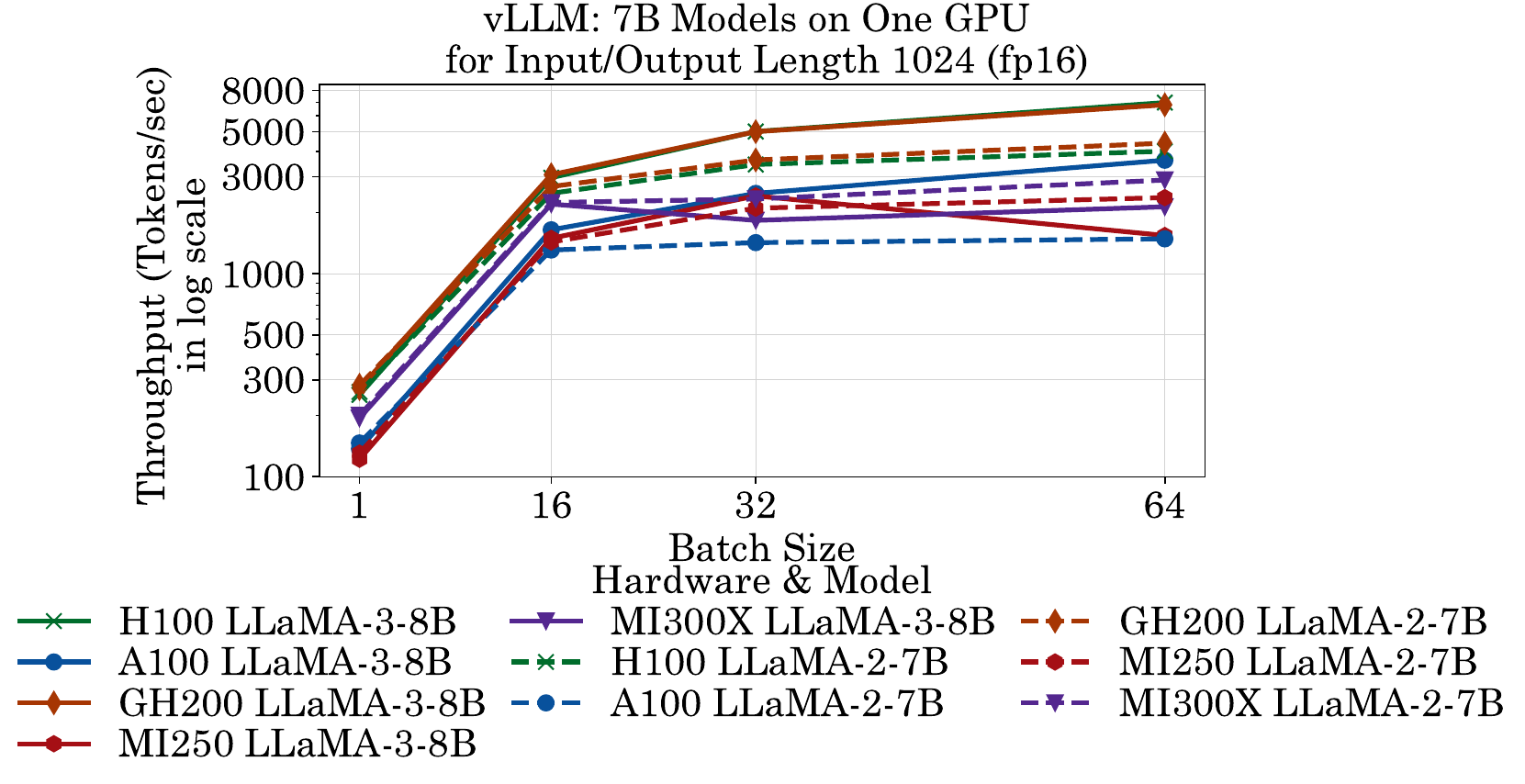}
        \caption{Throughput of 7B Models using vLLM 
        }
        \label{fig:vLLM_7B_across_hardware}
        \captionsetup{justification=centering}
\end{figure}

\begin{figure}
     \centering
        \includegraphics[width=\linewidth]{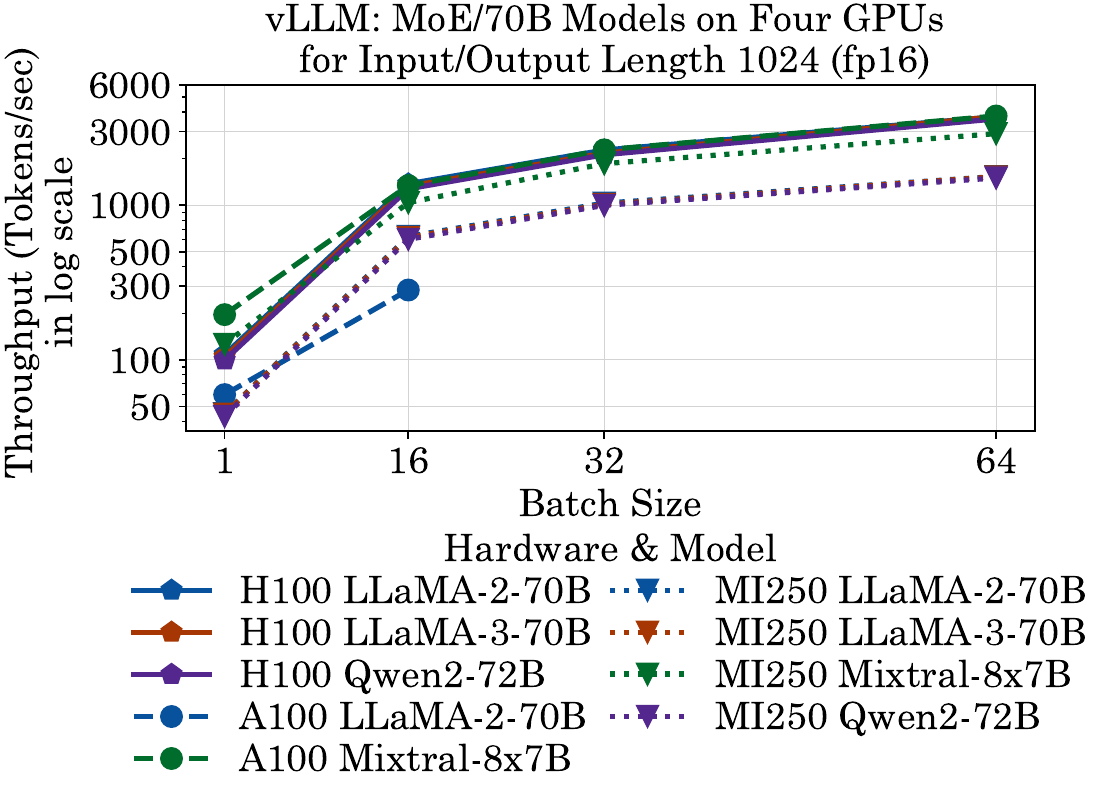}
        \caption{Throughput of 70B Models using vLLM}
        \label{fig:vLLM_70B_across_hardware}
        \captionsetup{justification=centering}
\end{figure}

Figure \ref{fig:7b_Perplexity_vs_Throughput_A100} compares the perplexity and throughput of $\sim$7B models, such as DeciLM \cite{decilm_7b}, GPT-J-6B \cite{brown2020language}, OPT-6.7B \cite{zhang2022opt}, Gemma-7B \cite{team2024gemma}, Qwen1.5-7B \cite{Qwen2_1_5b}, Aquila-7B \cite{aquilachat_7B} and Bloom-7.1B \cite{bloom_7b1}, on A100 and H100 respectively on LongBench dataset \cite{bai2023longbench}. LLaMA-2-7B has better perplexity than LLaMA-3-8B and Mistral-7B due to MHSA in LLaMA-2-7B over GQA in the latter two models. While GQA balances speed and performance, MHSA improves the model's validation performance. DeciLM-7B has the highest throughput, while Mistral-7B provides a good tradeoff with only 0.09 higher perplexity than LLaMA-2-7B and 0.8 times less throughput than DeciLM-7B. Gemma-7B has the lowest throughput, attributed to its larger head and intermediate size.

\begin{figure}[H]
    \centering
        \includegraphics[width=0.75\linewidth]{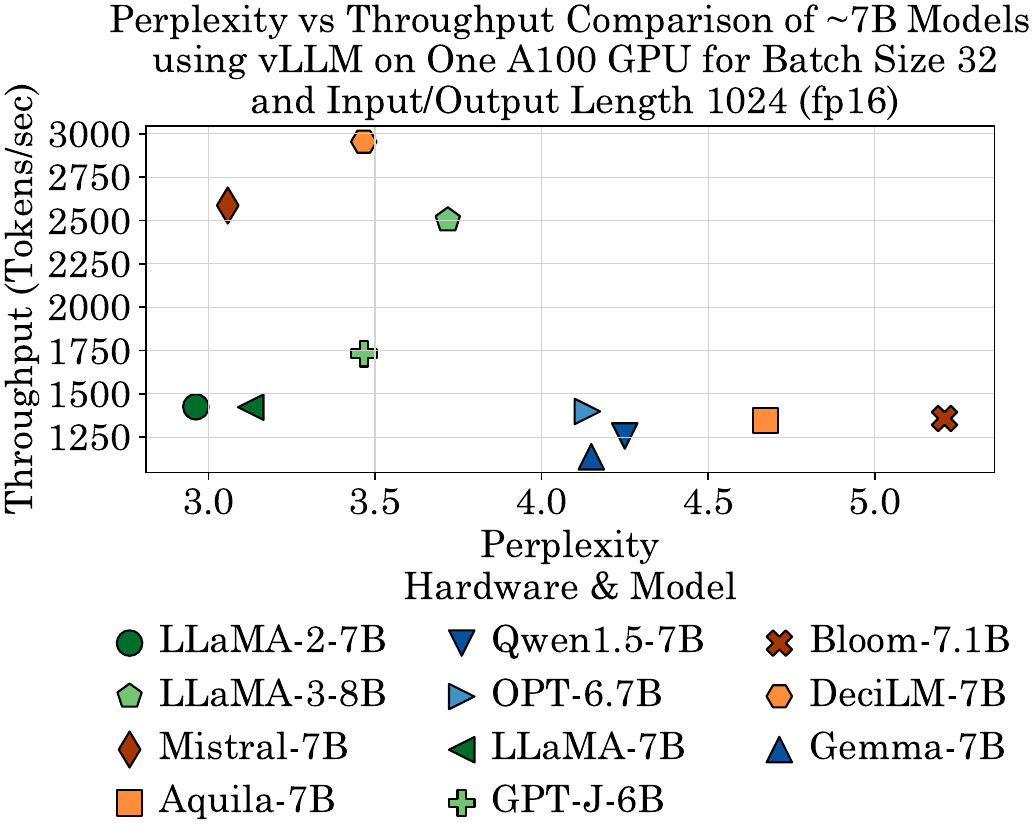}
        \caption{Perplexity vs A100 Throughput}
        \label{fig:7b_Perplexity_vs_Throughput_A100}
    \captionsetup{justification=centering}
\end{figure}

\subsubsection{\textbf{DeepSpeed-MII}} DS-MII is an easy-to-use framework that supports dynamic split fusion to combine multiple operations and CUDA compiler optimizations. Despite its advantages, the framework is limited to specific Nvidia GPUs and does not have an optimized implementation of efficient neural operators. It applies system optimizations based on the model type, batch size, and available hardware resources. While 7B models exhibit good scalability across 1, 2, and 4 devices with increasing batch size, performance discrepancies are observed compared to TRT-LLM and vLLM with respect to neural architecture. Figure \ref{fig:ds_mii_scaling} illustrates that LLaMA-2-7B (MHSA) using DS-MII outperforms LLaMA-3-8B (GQA), and LLaMA-3-8B surpasses Mistral-7B, contrary to the expectation that GQA runs faster than MHSA. On a single A100 GPU, LLaMA-2-7B is 1.18 times faster than LLaMA-3-8B for a batch size of 64 and input/output length of 128. However, DS-MII is particularly useful for big models and large batch sizes which is illustrated in Figure \ref{fig:TRT_DS_vLLM}. For the Mixtral-8x7b model, DS-MII outperforms vLLM for relatively large batch sizes and sequence lengths. DS-MII is 1.04x faster than vLLM for batch size 64 and Input/output length 2048.

\begin{figure}
    \centering
    \includegraphics[width=0.8\linewidth]{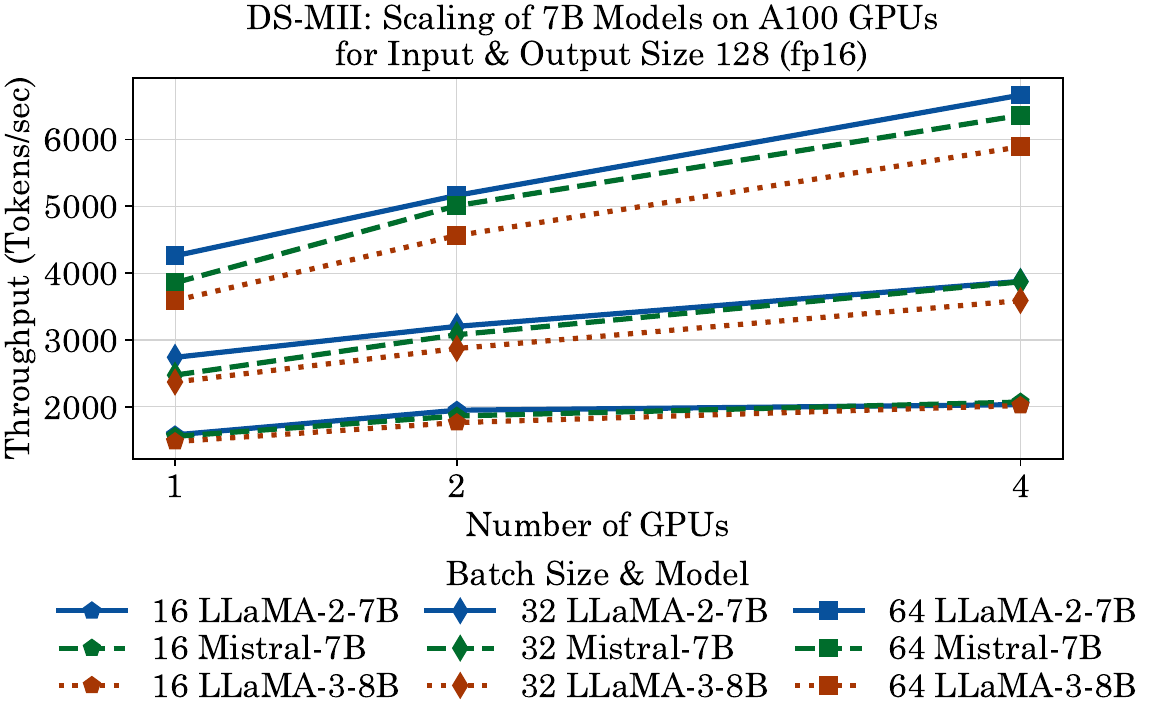}
    \caption{7B Models using DS-MII on A100 GPUs}
    \label{fig:ds_mii_scaling}
    \captionsetup{justification=centering}
\end{figure}

\begin{figure}[H]
    \centering
    \includegraphics[width=0.8\linewidth]{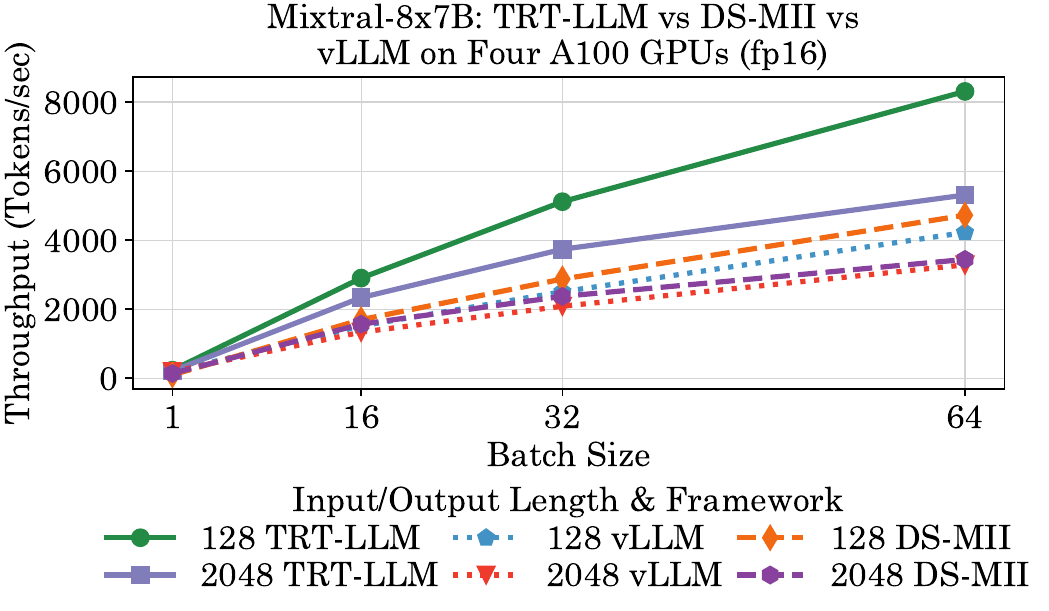}
    \caption{Mixtral-8x7B Comparison on A100 GPU}
    \label{fig:TRT_DS_vLLM}
    \captionsetup{justification=centering}
\end{figure}


\subsubsection{\textbf{llama.cpp}} llama.cpp is designed to perform LLM inference with minimal setup and accelerate on a wide variety of hardware. Although llama.cpp can be integrated seamlessly across devices, it suffers from device scaling across AMD and Nvidia platforms batch sizes due to the inability to fully utilize parallelism and LLM optimizations. 
Figure \ref{fig:7B_llama_cpp_batch} show llama.cpp's marginal performance benefits with an increase in GPU count across diverse platforms. 

\begin{figure}[H]
\centering
\includegraphics[width=0.90\linewidth]{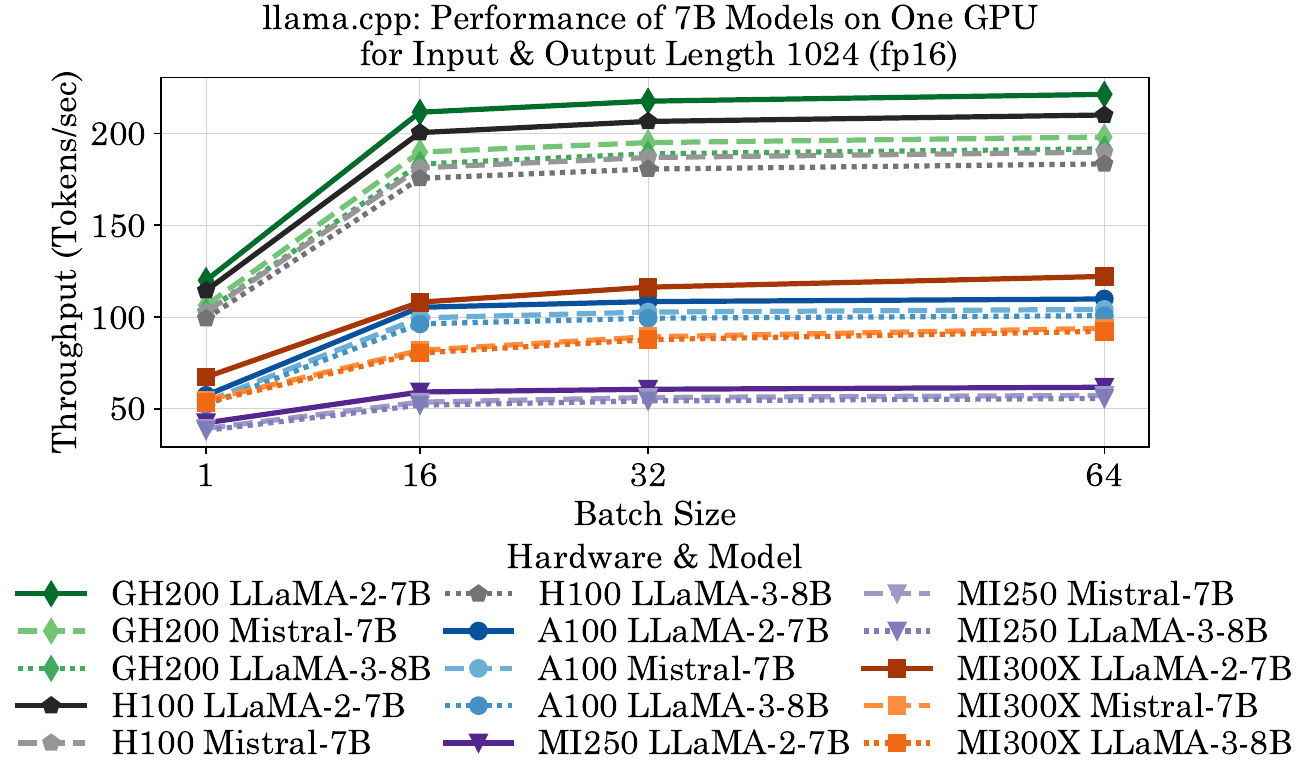}
\caption{Throughput of 7B Models using llama.cpp}
\label{fig:7B_llama_cpp_batch}
\captionsetup{justification=centering}
\end{figure}

Figure \ref{fig:llama_cpp_scaling} demonstrated weak scaling of llama.cpp. Also, LLaMA-2-7B outperforms both Mistral-7B and LLaMA-3-8B, while Mistral-7B surpasses LLaMA-3-8B across different batch sizes and number of GPUs. This is counterintuitive compared to TRT-LLM and vLLM, where models with GQA perform better than MHSA. This shows that llama.cpp is unable to fully take the advantage of Group Query Attention.

\begin{figure}[H]
\centering
\includegraphics[width=0.9\linewidth]{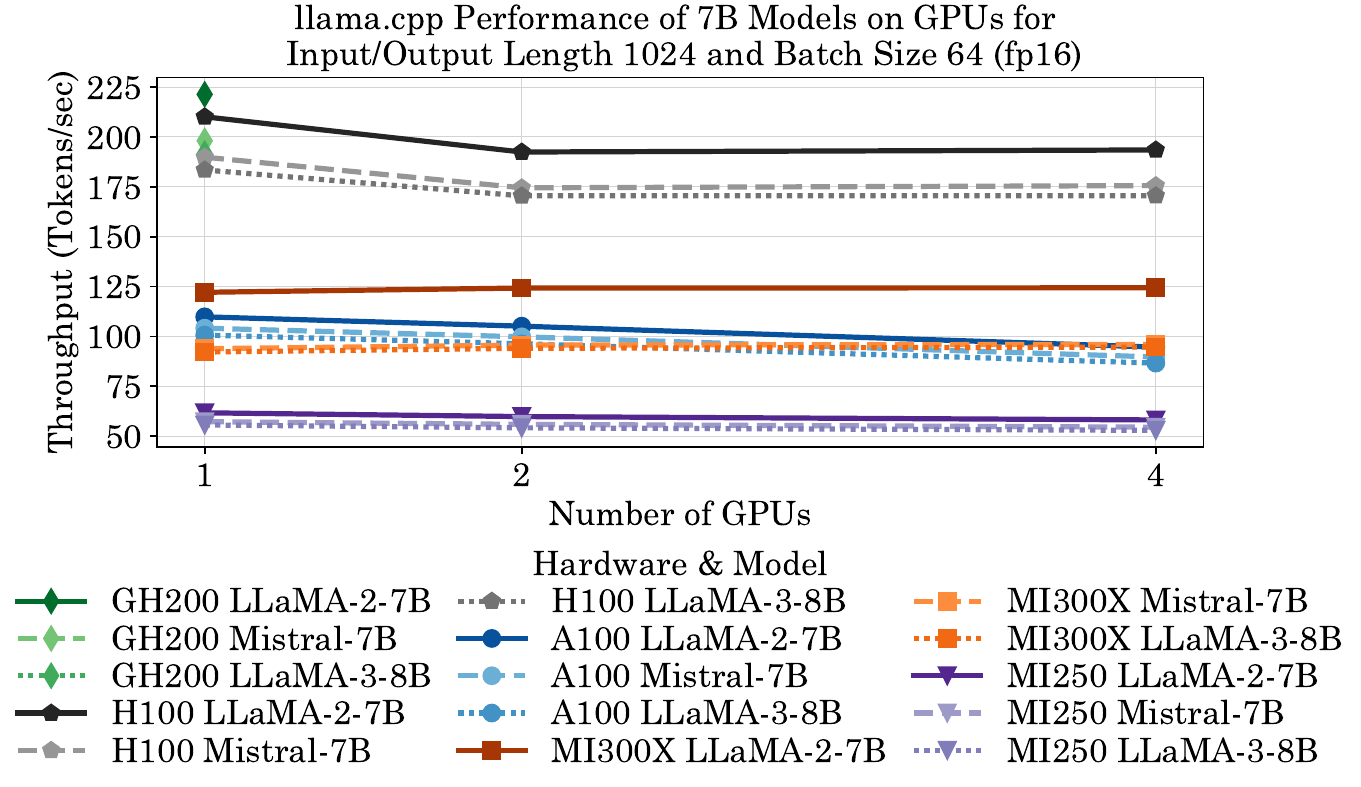}
\caption{llama.cpp: 7B Model Scaling}
\label{fig:llama_cpp_scaling}
\captionsetup{justification=centering}
\end{figure}


\section{Hardware-wise Benchmarking}

This section presents a comprehensive analysis of the performance characteristics of LLMs across a wide range of AI accelerators. We aim to offer insights into the strengths and limitations of each hardware platform for LLMs. 

\subsubsection{\textbf{Nvidia GPUs}} 
Figure \ref{fig:A100_framework_compare_7B} compares 7B models on A100 using different frameworks. TRT-LLM outperforms vLLM and DS-MII on Nvidia hardware by leveraging TensorRT, CUDA, and cuDNN to optimize matrix computations. It employs advanced techniques like layer fusion, kernel auto-tuning, and dynamic tensor memory management to reduce latency and memory footprint. These optimizations improve scalability, making TensorRT-LLM well-suited for high-performance inference on Nvidia GPUs. Also, we observe that llama.cpp is the slowest of the frameworks due to suboptimal usage of device optimizations. Llama.cpp lacks full implementation of tensor parallelism and does not leverage the full potential of Tensor Cores, leading to the underutilization of GPU. From a model perspective, Mistral-7B performs better than LLaMA-3-8B. Both the models share the same architecture and configuration, including hidden size, number of layers, and FFN size, with the primary difference being that LLaMA-3-8B has a vocab size four times larger than Mistral-7B, causing the later model to achieve more throughput.

\begin{figure}
    \centering
    \includegraphics[width=0.9\linewidth]{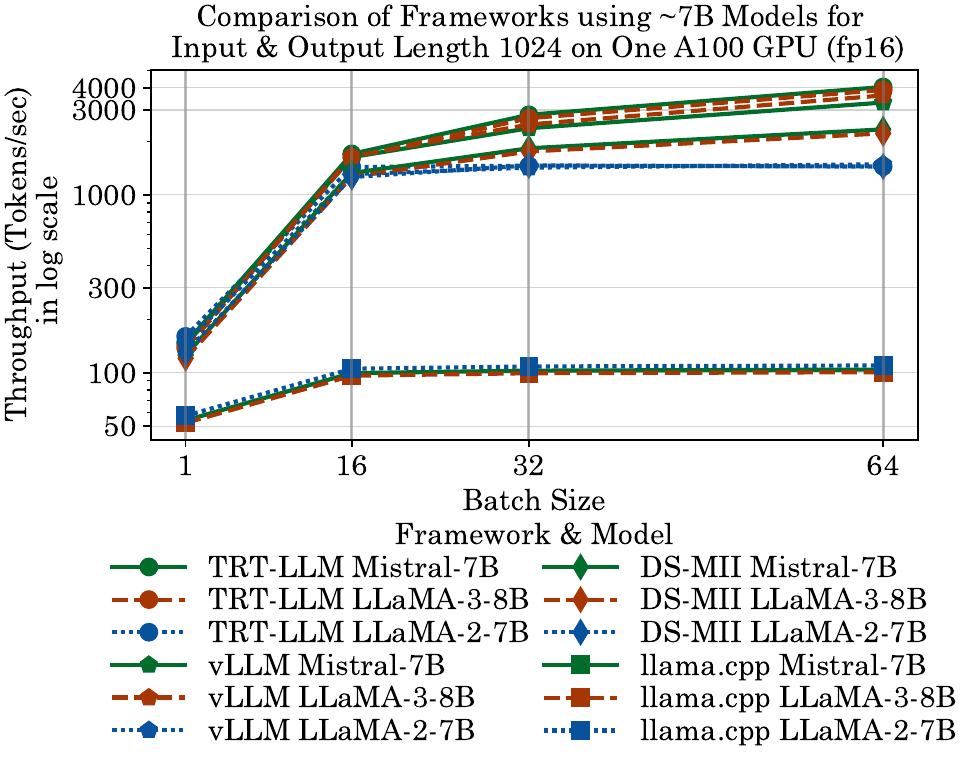}
    \captionsetup{justification=centering}
    \caption{Throughput of 7B Models on A100}
    \label{fig:A100_framework_compare_7B}
\end{figure}

Figure \ref{fig:7B_Power}(left) illustrates the power consumption, and Figure \ref{fig:7B_Power}(right) illustrates throughput per watt of the LLaMA-2-7B and LLaMA-3-8B models on A100, H100 and GH200 using vLLM and TRT-LLM frameworks. LLaMA-2-7B on GH200 using TRT-LLM consumes more power than A100 and H100, while LLaMA-3-8B on A100 consumes the lowest. The performance per watt ratio for LLaMA-3-8B across all frameworks and hardware is higher than LLaMA-2-7B for the same hardware and software setting. TRT-LLM consumes more power than vLLM due to more utilization of the hardware and delivers more performance per watt.




\begin{figure}
    \centering
    \includegraphics[width=\linewidth]{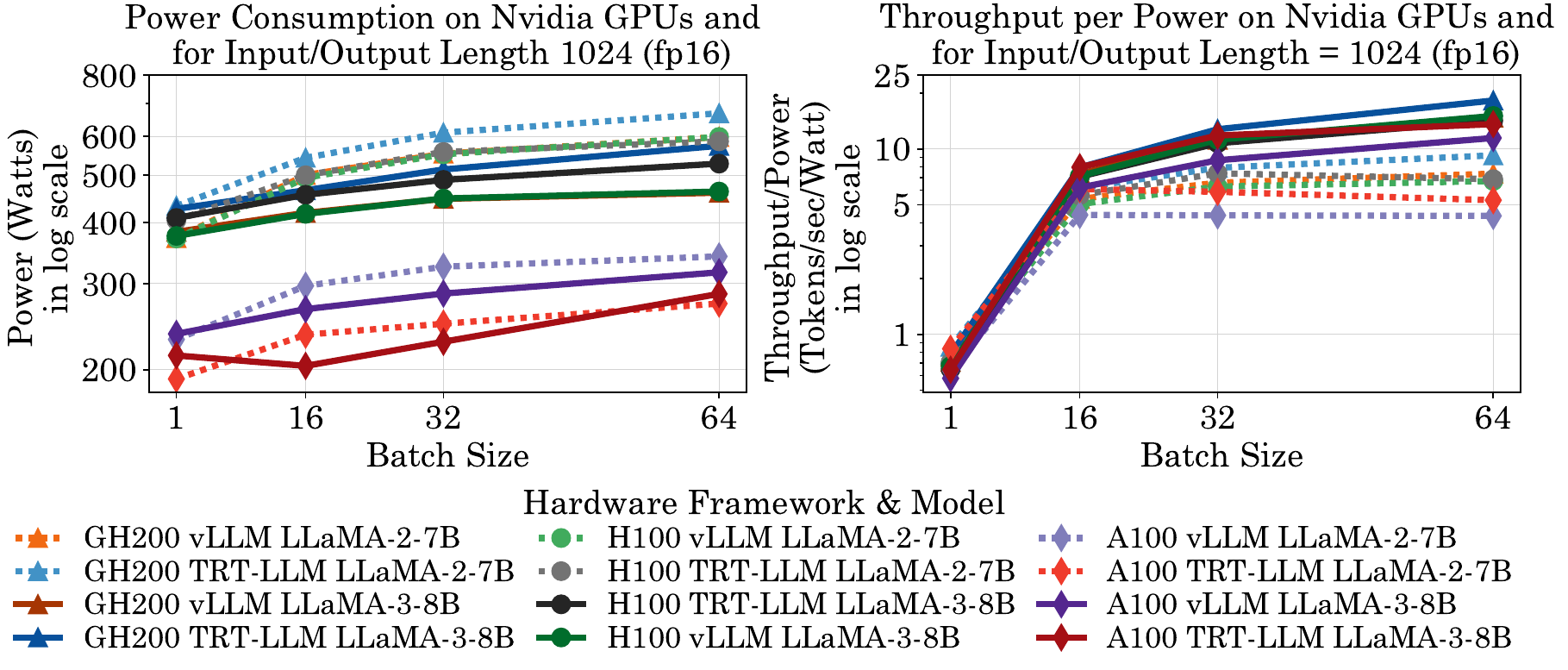}
    \captionsetup{justification=centering}
    \caption{Power Consumption and Throughput per Watt}
    \label{fig:7B_Power}
\end{figure}

\subsubsection{\textbf{AMD MI250 GPU}} AMD GPUs leveraging ROCm support frameworks like vLLM and llama.cpp and can offer performance comparable to A100 for specific scenarios. Figure \ref{fig:LLaMA_3_8B_MI250} highlights a notable trend in the MI250 GPU's performance. Compared to the A100, the MI250's compute and memory units reach saturation more rapidly. This is because we used Non-uniform memory access balancing, which forces the GPU to wait for the preemptive memory management unit notifier to derive page faults. The throughput of LLaMA-3-8B drops beyond batch size 32 with an increase in input/output length for the same batch size. The throughput trend will likely increase with increasing number of GPUs.

\begin{figure}[H]
\centering
\includegraphics[width=0.9\linewidth]{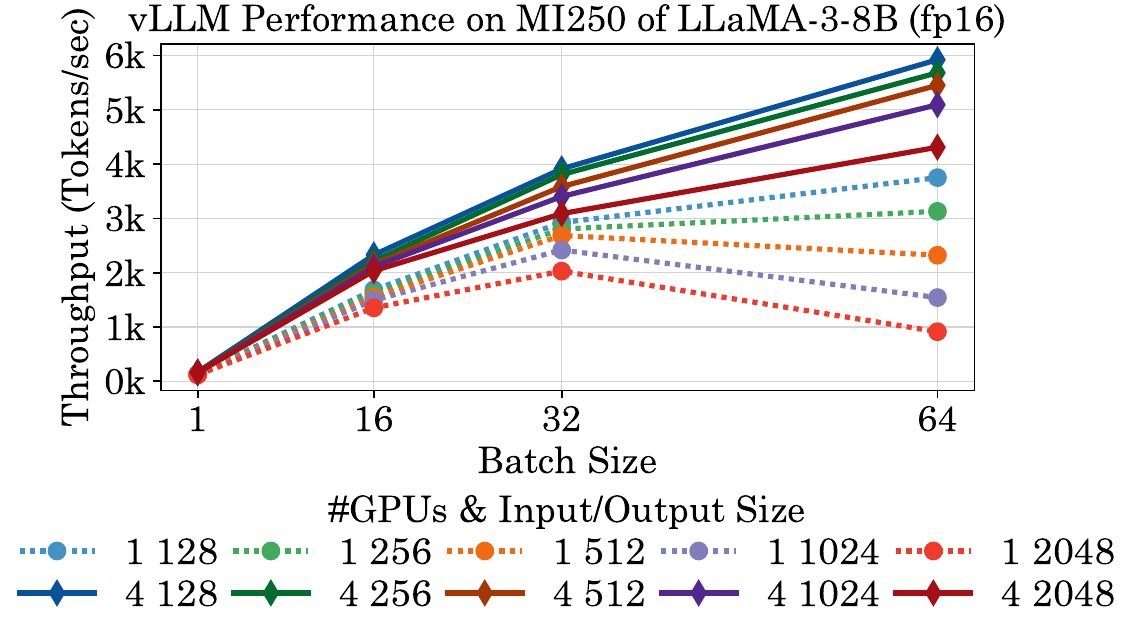}
\caption{LLaMA-3-8B using vLLM on single MI250 GPU}
\label{fig:LLaMA_3_8B_MI250}
\captionsetup{justification=centering}
\end{figure}

\subsubsection{\textbf{SambaNova SN40L}} Figures \ref{fig:SN40L_7B} and \ref{fig:SN40L_70B} illustrate the performance comparison of 8 SN40L RDUs (TP = 8) with A100 for 7B and 70B Models, which better performance than H100 and A100. Throughput increases with increasing input/output length (till 512) as SN40L handles short and long sequences differently. This contradicts the earlier observation that throughput decreases with an increase in input/output length. LLaMA-3-8B and Mistral-7B outperform LLaMA-2-7B on SN40L as the compiler improvements for small-sized models were not applied to the LLaMA-2-7B model compared to other LLMs. 
The accelerator has a 3-tier memory system unlike the traditional 2-tier memory system in GPUs.

\begin{figure}
\centering
\includegraphics[width=\linewidth]{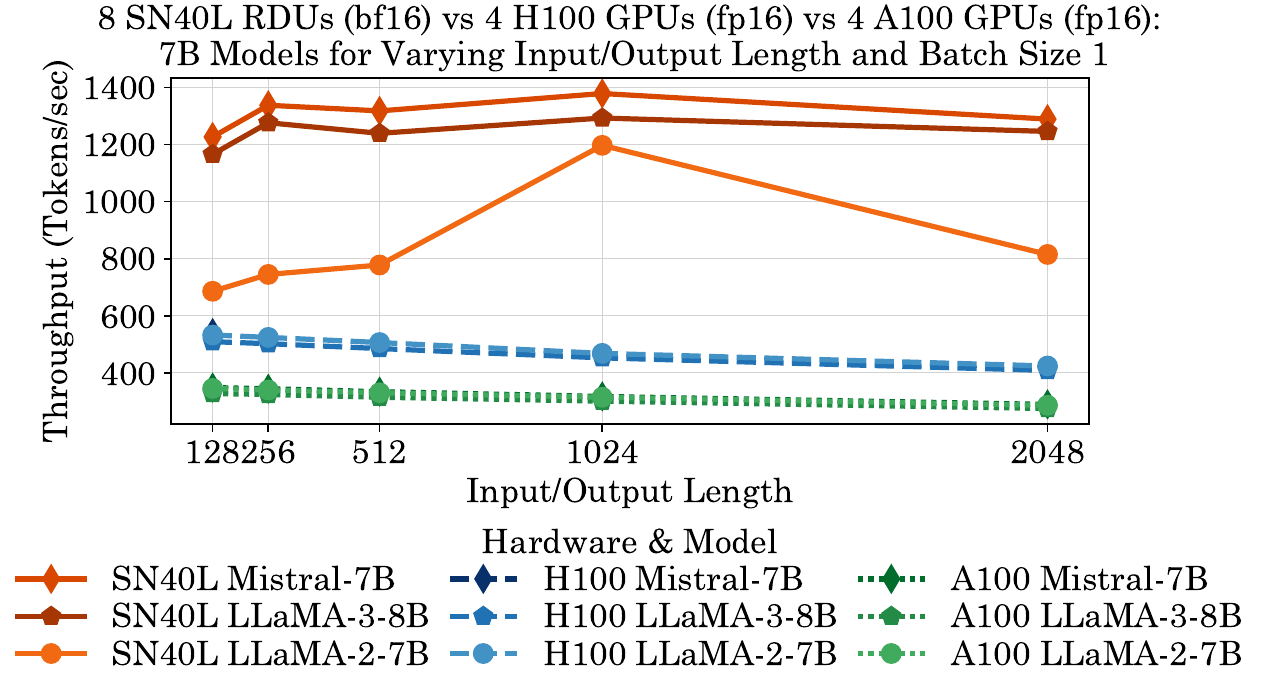}
\caption{Throughput Comparison of 7B Models on 8 SN40L RDUs with 4 H100s and 4 A100s GPU}
\label{fig:SN40L_7B}
\captionsetup{justification=centering}
\end{figure}

\begin{figure}
\centering
\includegraphics[width=\linewidth]{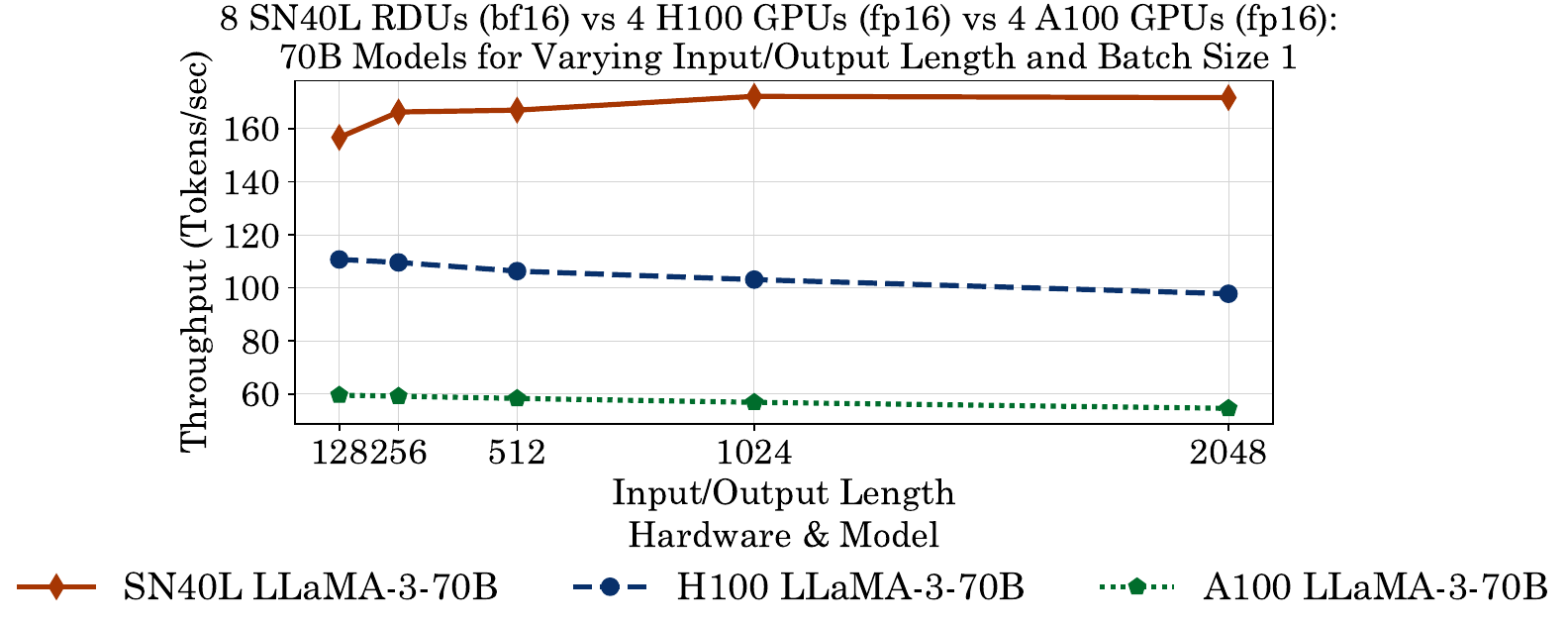}
\caption{Throughput Comparison of 70B Model on 8 SN40L RDUs with 4 A100 and 4 H100 GPUs}
\label{fig:SN40L_70B}
\captionsetup{justification=centering}
\end{figure}

\subsubsection{\textbf{Habana Gaudi2}} Figure \ref{fig:7B_Gaudi2} and \ref{fig:70B_Gaudi2} compare the 7B models performance on Gaudi2, H100 and A100. The throughput of Gaudi2 is better than A100 due to parallel operations and efficient matrix multiplication. Gaudi2's heterogeneous architecture allows for overlapping compute time between its matrix multiplication engine and tensor processing core (TPC), while A100 lack this parallel execution capability. Also, Gaudi2 uses multiple smaller matrix accelerators instead of a single large one, which requires less bandwidth to fully utilize its compute power. However, our experiments showed that Habana Gaudi2 attains memory issues quicker than other accelerators for the same model configuration. 

\begin{figure}
\centering
\includegraphics[width=\linewidth]{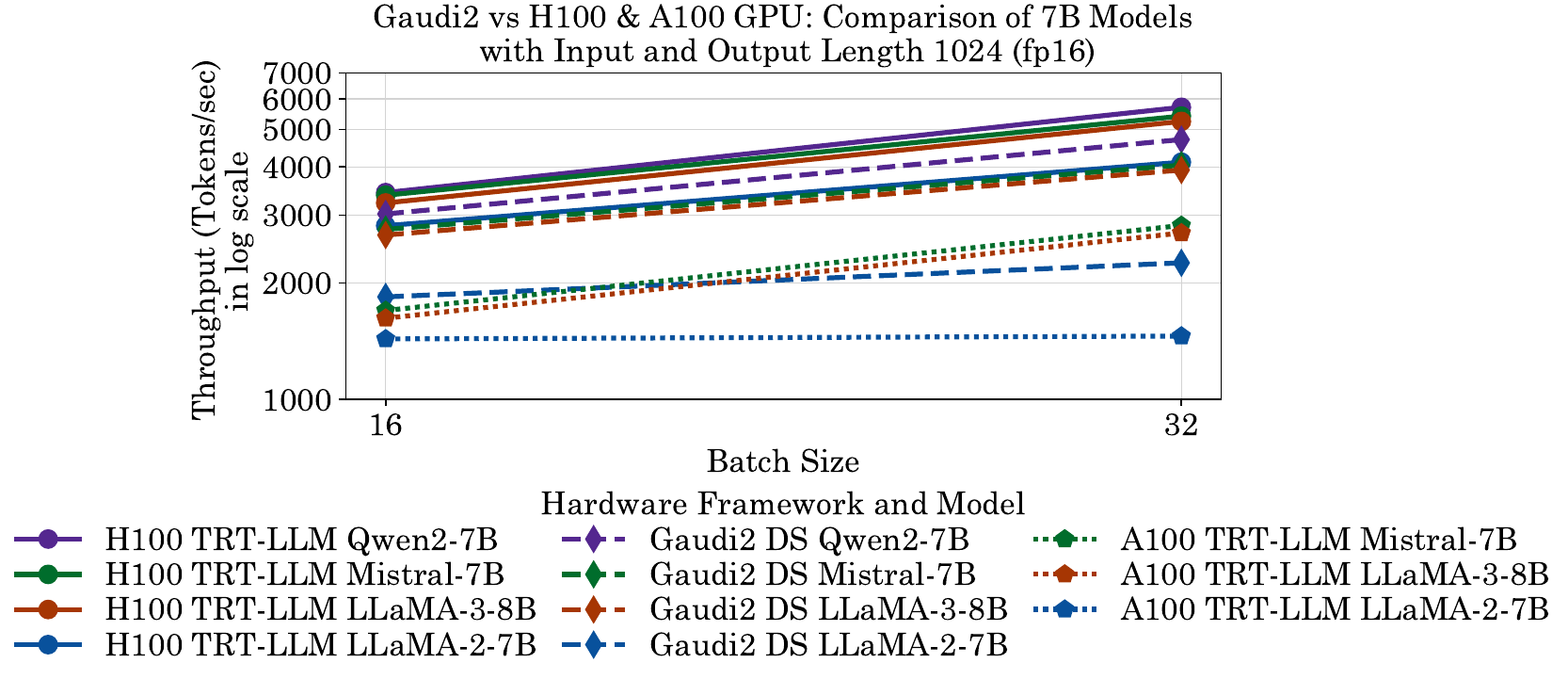}
\caption{H100 vs A100 vs Gaudi2: 7B Models}
\label{fig:7B_Gaudi2}
\captionsetup{justification=centering}
\end{figure}



\section{Insights and Discussion}
\label{sec:insights}

In this section, we summarize the key findings of the analysis from three perspectives: framework-wise, accelerator-wise, and LLM architecture-wise. The core insights focus on GQA support, efficient KV cache management, enhanced utilization of computing kernels, and improved accessibility of both hardware and software.

\subsubsection{\textbf{Framework-wise Takeaways}} Several factors must be considered when choosing an LLM inference framework for large-scale deployment. 
An ideal framework should deliver \textit{high throughput and low latency}. It should \textit{scale} well with an increase in batch sizes as memory requirements grow linearly, necessitating efficient management, especially concerning the KV Cache. 
For instance, Llama.cpp is highly portable and can be accelerated across various software stacks and hardware platforms with minimal code changes. However, it experiences weak scaling and does not significantly improve for large batch sizes as the framework does not utilize compute resources well. 
DS-MII scales very well with an increase in GPUs, batch sizes and sequence lengths and integrates seamlessly with PyTorch, but it is limited to Nvidia GPUs.
The framework should \textit{support diverse hardware accelerators} to democratize LLMs and scale with an increasing number of computing chips. For instance, TensorRT-LLM on Nvidia GPUs offers the highest performance but is limited to specific platforms. In contrast, vLLM supports a broader range of hardware but consumes more power and is slower than TensorRT-LLM on Nvidia GPUs.
A framework should be \textit{flexible} and adaptable to new model architectures, enabling quick deployment of new models. For instance, DS-MII and vLLM are relatively easy to use and can directly use Huggingface (HF) model weights. However, TRT-LLM and llama.cpp requires users to convert to HF weights to Tensor-RT engine and GGUF format which requires customizing to individual models and operations. 
Frameworks should \textit{leverage optimizations} like efficient attention mechanisms such as GQA. For example, LLaMA-3-8B and Mistral-7B outperform LLaMA-2-7B with TensorRT-LLM and vLLM, whereas LLaMA-3-8B cannot perform better than LLaMA-2-7B with llama.cpp and Deepspeed-MII as they do not support model-wise optimizations well. GQA models are technically superior to MHSA in terms of performance. Overall, the choice of framework should be tailored to specific user scenarios and infrastructure constraints.

\begin{figure}
        \centering
        \includegraphics[width=0.95\linewidth]{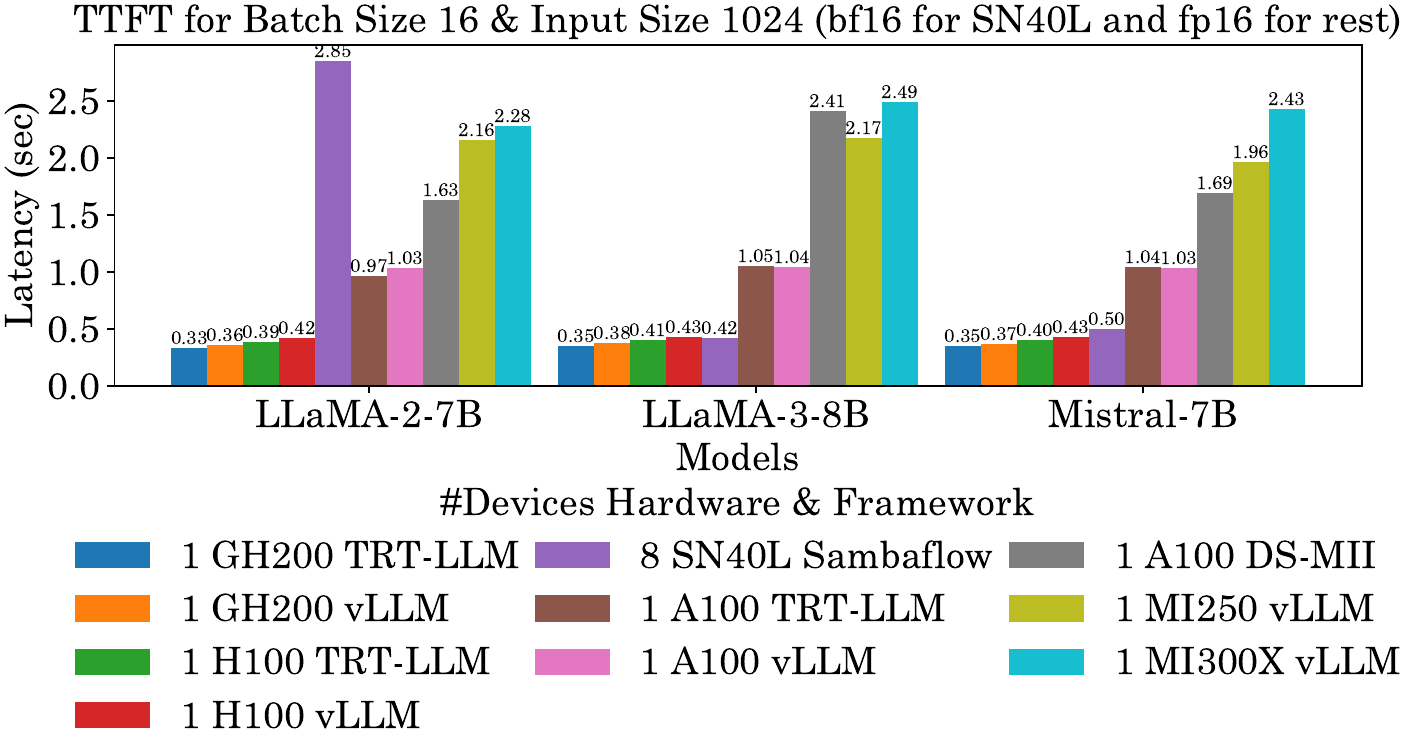}
        \caption{Time to First Token (TTFT)}
        \label{fig:TTFT}
        \captionsetup{justification=centering}
\end{figure}

\begin{figure}
        \centering
        \includegraphics[width=0.95\linewidth]{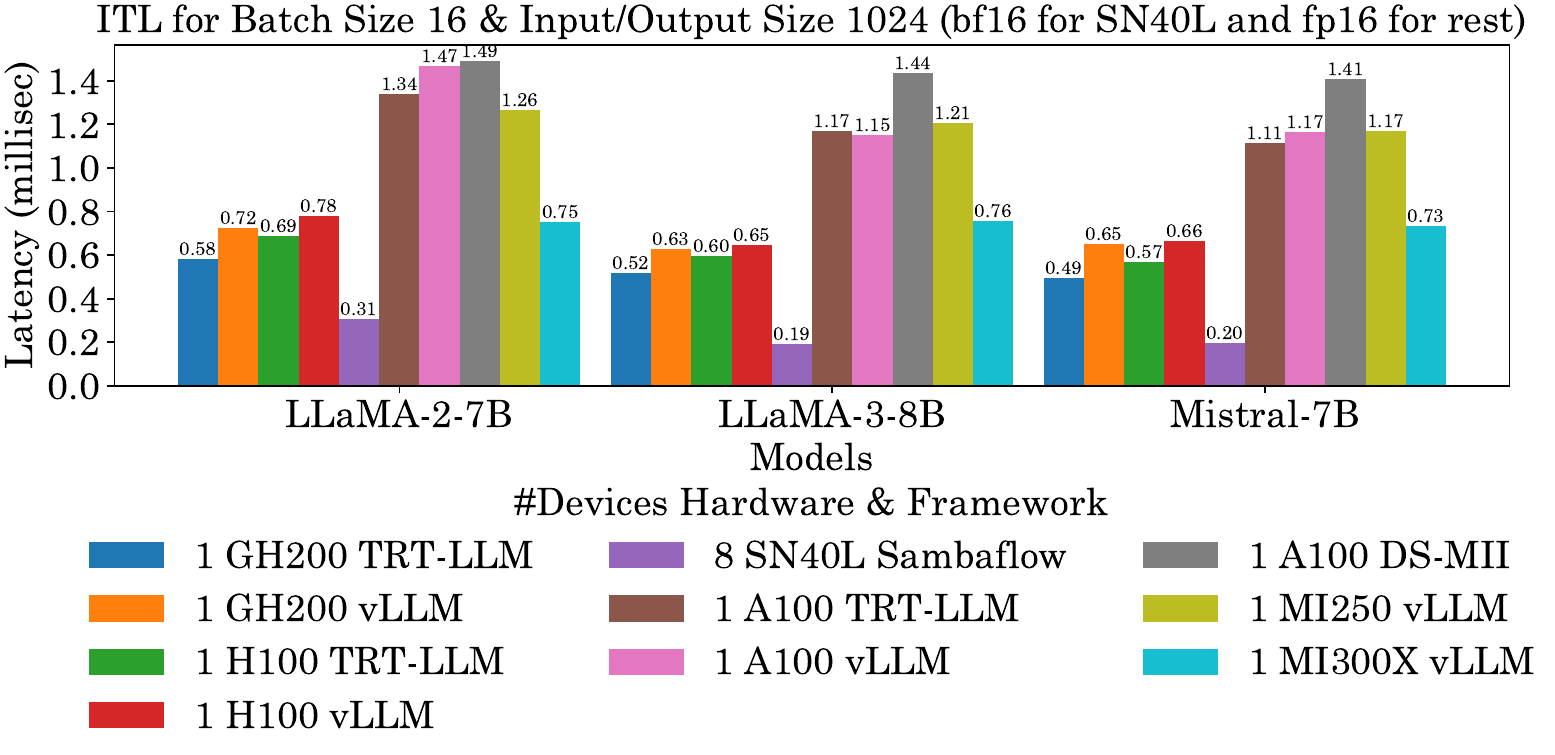}
        \caption{Inter Token Latency (ITL)}
        \label{fig:ITL}
        \captionsetup{justification=centering}
\end{figure}


\subsubsection{\textbf{Accelerator-wise Takeaways}} The choice of AI accelerators for LLM inference depends on several factors, such as availability, large-scale acceleration, power consumption, and scaling of inference frameworks. 
Nvidia GPUs are widely available with optimizations from the hardware and software end. However, the power consumption of hardware such as H100 and return for power investment is better for lower versions of Nvidia GPU such as A100. 
An accelerator should support a wide range of inference frameworks and be capable of taking advantage of SOTA optimizations. For instance, MI250 GPUs are comparable to A100 in running vLLM for certain scenarios. However, it suffers from early saturation where the performance drops beyond a batch size and token length faster than A100 GPU. On the other hand, Gaudi2 outperforms A100 but faces Out-of-memory issues for large batch sizes and does not support a wide range of frameworks. 
The current SN40L setup is limited to serving only a few batch sizes and a fixed number of RDUs (8 in our case). It is available as an online inference service Sambastudio \cite{samba_studio}, and support for wider LLMs is not yet supported, unlike GPUs. 
Figures \ref{fig:7B_across_HW_Batch_size} and \ref{fig:7B_across_HW_input_output} compare LLaMA-3-8B model across different hardware for varying batch sizes and input/output lengths. We observe that SN40L has the best performance up to batch size 32. The chat-based applications prioritize the rapid display of output tokens to enable immediate reading, making Time to First Token (TTFT) crucial. As new tokens are generated and read by users, Inter Token Latency (ITL) becomes increasingly significant to maintain a smooth conversational flow. Figures \ref{fig:TTFT} and \ref{fig:ITL} depict the Time to First Token (TTFT) and Inter-Token Latency (ITL) for LLaMA-2-7B, LLaMA-3-8B, and Mistral-7B models. The results reveal that while SN40L exhibits higher TTFT compared to other hardware, it demonstrates lower ITL, indicating faster token generation after the initial output. Figure \ref{fig:7B_peak_perf} illustrates the peak performance of 7B models on different hardware platforms. Overall, each accelerator presents unique trade-offs in terms of performance, accessibility, and compatibility with various frameworks, which must be considered for LLM deployment.

\begin{figure}[H]
        \centering
        \includegraphics[width=0.9\linewidth]{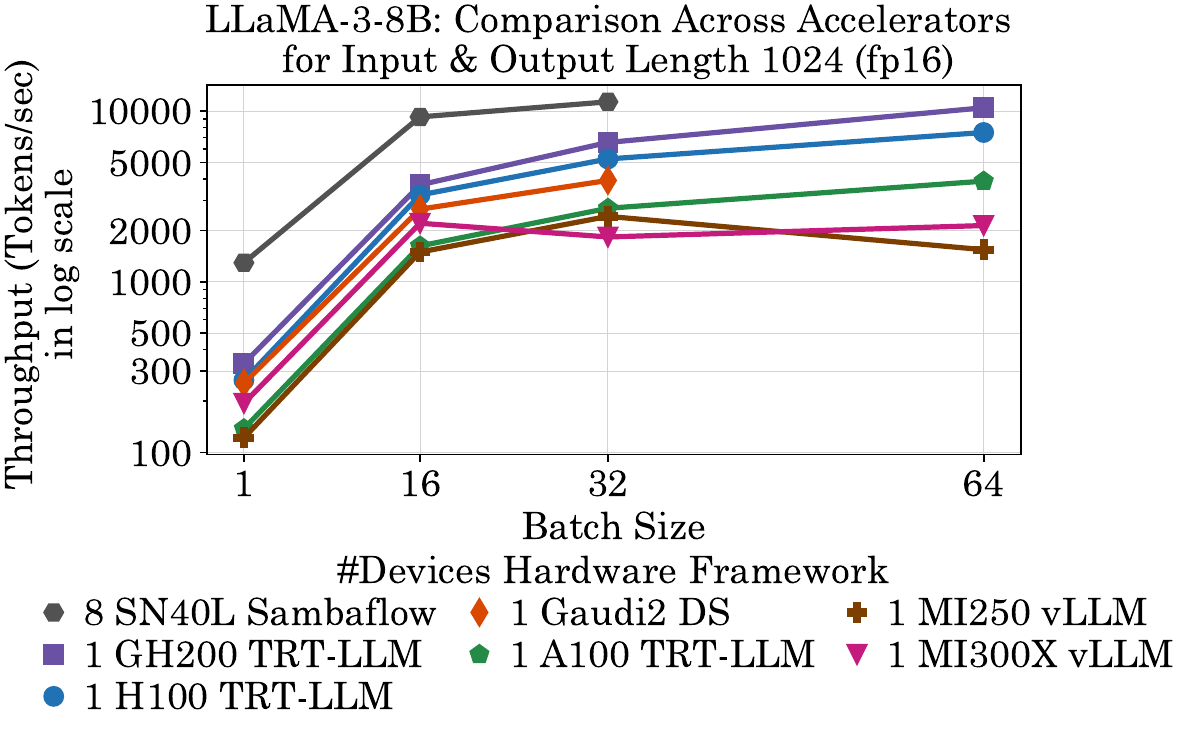}
        \caption{Throughput vs Batch Size}
        \label{fig:7B_across_HW_Batch_size}
        \captionsetup{justification=centering}
\end{figure}

\vspace{-4mm}

\begin{figure}[H]
        \centering
        \includegraphics[width=0.9\linewidth]{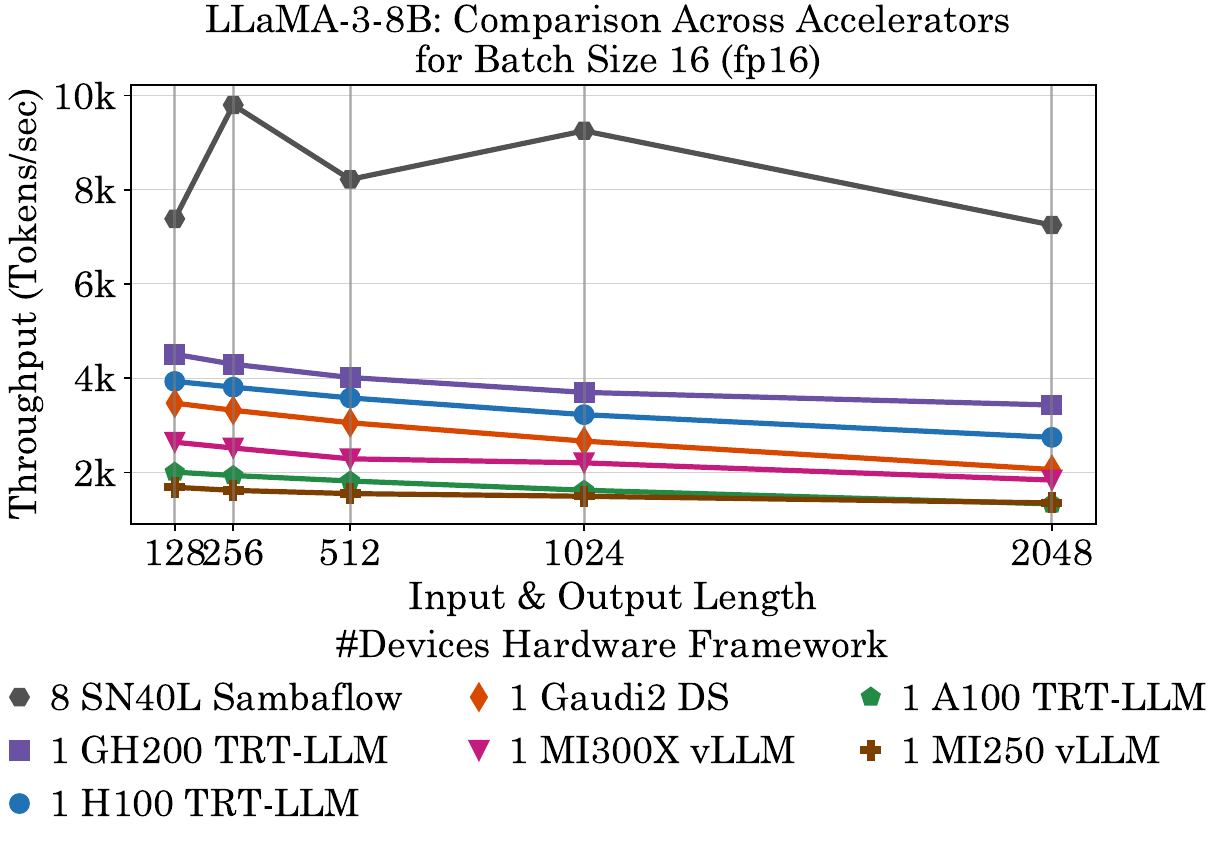}
        \caption{Throughput vs Input/Output Length}
        \label{fig:7B_across_HW_input_output}
        \captionsetup{justification=centering}
\end{figure}

\subsubsection{\textbf{LLM Architecture-wise Takeaways}} Model performance varies significantly with size and architecture. Smaller models, like a 7B model, typically offer better throughput than larger ones, such as a 70B model. However, within similar size ranges, performance differences can be attributed to operation types, hyperparameters, and batch sizes. For instance, LLaMA-2-7B, with a smaller FFN size and larger attention size (MHSA) compared to LLaMA-3-8B and Mistral-7B (GQA), performs well at lower batch sizes but declines as batch size increases due to attention and KV cache demands. Conversely, the Qwen2-7B model, with a larger vocabulary and fewer hyperparameters, outperforms others because the vocabulary size primarily affects inputs and outputs, leaving the core model with fewer parameters. Among 70B models, LLaMA-2-70B is slightly more efficient across accelerators due to its smaller vocabulary compared to LLaMA-3-70B and Qwen2-72B. The Mixtral-7x8B MoE model surpasses 70B models by activating only two experts per layer during inference, effectively functioning as a 14B model. Our findings consistently demonstrate that platforms, frameworks, and accelerators lacking advanced optimization techniques exhibit inferior performance. Specifically, Deepspeed-MII and llama.cpp, which underutilize GQA optimization, are outperformed by TensorRT-LLM and vLLM. Figures \ref{fig:TTFT} and \ref{fig:ITL} show that LLaMA-2-7B relatively requires less time to generate the first token (could be attributed to small FNN dimension compared to Mistral-7B and LLaMA-3-8B as the first token does not significantly high KV cache memory). However, ITL is high compared to Mistral-7B and LLaMA-3-8B, resulting in less throughput on different systems.


\begin{figure}[H]
        \centering
        \includegraphics[width=0.9\linewidth]{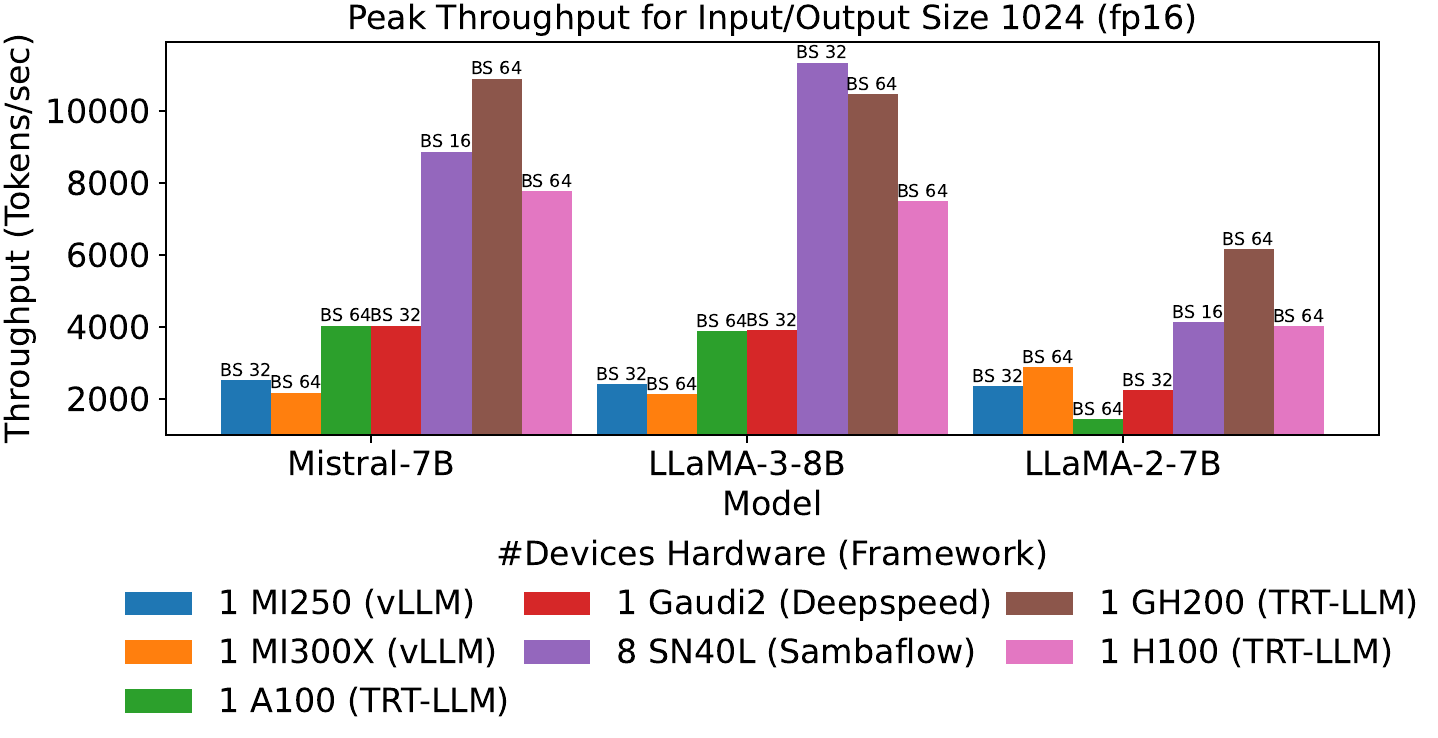}
         \caption{Peak Performance\protect\footnotemark}
        \label{fig:7B_peak_perf}
        \captionsetup{justification=centering}
\end{figure}
\vspace{-4mm}

\footnotetext{The peak performance mentioned here is throughput in our benchmark study. NVIDIA GPUs and SN40L can handle batch sizes beyond 32 and 64, respectively, and therefore, peak throughput might be higher. Conversely, the performance of AMD GPUs declines beyond a certain batch size. We encountered out-of-memory issues on Gaudi2 at batch sizes of 32 and 64 in several test scenarios. The paper's MI250, MI300X and Gaudi2 numbers are out-of-the-box without special optimization flags.}

\section{Conclusion}

We introduce \textit{LLM-Inference-Bench}, a comprehensive benchmarking suite that evaluates the inference performance of the variety of llama-style LLMs across SOTA AI accelerators using widely available LLM inference frameworks. We provide insights about different models and their behavior across different frameworks and accelerators. TensorRT-LLM provides the highest performance and lowest power consumption on Nvidia platforms, while vLLM can be accelerated on a variety of devices. While dealing with AI accelerators, vendor-specific frameworks result in the best throughput. Nvidia H100 GPU offers the best throughput and performance per watt across all GPUs. 
Mistral-7B offers the best throughput and perplexity tradeoff compared to several SOTA 7B models. In addition to the outcomes discussed in this paper, we also created a web-based interactive dashboard that can be used by AI researchers to determine the optimal configuration of framework, accelerator and model suited for their workload.




\section*{Acknowledgements}
This research used resources of the Argonne Leadership Computing Facility, a U.S. Department of Energy (DOE) Office of Science user facility at Argonne National Laboratory and is based on research supported by the U.S. DOE Office of Science-Advanced Scientific Computing Research Program, under Contract No. DE-AC02-06CH11357.
We gratefully acknowledge the computing resources provided and operated by the Joint Laboratory for System Evaluation (JLSE) at Argonne National Laboratory. We would like to thank Ashutosh Dhar and Geetika Gupta from Nvidia; Darshan Gandhi, Dawei Huang, Jennifer Glore, Sankar Rachuru and Connor McCormick from SambaNova and Chenna Bayapureddy and Yuting Yang from Intel Habana for their valuable feedback.

\newpage
\clearpage

\bibliographystyle{IEEEtran}
\bibliography{bibliography/AI_Accelerators,
bibliography/github_repos,
bibliography/Inference_frameworks,
bibliography/LLM_HF_repos, 
bibliography/LLMs,
bibliography/Validation,
bibliography/survey}

\appendix
\appendices 

\section{LLM Primitives} 
\label{apx:llm_premitives}

\subsection{LLM Architectures} \label{appendix:llm_arch}

\subsubsection{Dense vs MoE model}

Dense and Mixture-of-Experts (MoE) models represent two distinct approaches, each with its own advantages and trade-offs. Dense models, characterized by FC layers where all parameters are used for every input, offer simplicity and ease of training but can become computationally prohibitive at extremely large scales. Notable examples of dense LLMs include LLaMA-2-7B \cite{touvron2023llama} and LLaMA-3-8B \cite{llama3_8b}. MoE models \cite{shazeer2017outrageously} employ a combination of specialized sub-networks or experts and a gating mechanism to selectively activate only a subset of parameters for each input, allowing for greater computational efficiency. While MoE models can achieve comparable or superior performance to dense models with similar computational costs, they typically require 2-4 times more total parameters. This increased parameter count results in higher memory requirements, making MoE models less efficient in I/O-bounded scenarios like autoregressive generation. MoE models, such as Mixtral-8x7B \cite{mixtral_8_7b} and Qwen2-57B-A14B \cite{Qwen2_57B_A14B}, implement multiple experts within the MLP block and the attention can be either MHSA or GQA. The attention module maintains a more conventional MHSA or GQA structure.  The variants of MoE architecture include Hybrid MoE \cite{snowflake_arctic} Hybrid Transformer-Mamba MoE \cite{lieber2024jamba}, and Composition of Experts (CoE) \cite{prabhakar2024sambanova}. Hybrid MoE combines elements of both MoE and dense models by integrating a residual MoE with a dense transformer. Jamba \cite{lieber2024jamba} presents a hybrid Transformer-Mamba MoE, which interleaves blocks of Transformer and Mamba layers \cite{gu2023mamba}, incorporating MoE in some layers. CoE represents a novel approach to MoE architectures by combining expert LLM networks to achieve improved performance or efficiency over individual models.

\begin{figure}
    \centering
    \includegraphics[width=7cm, height=6cm]{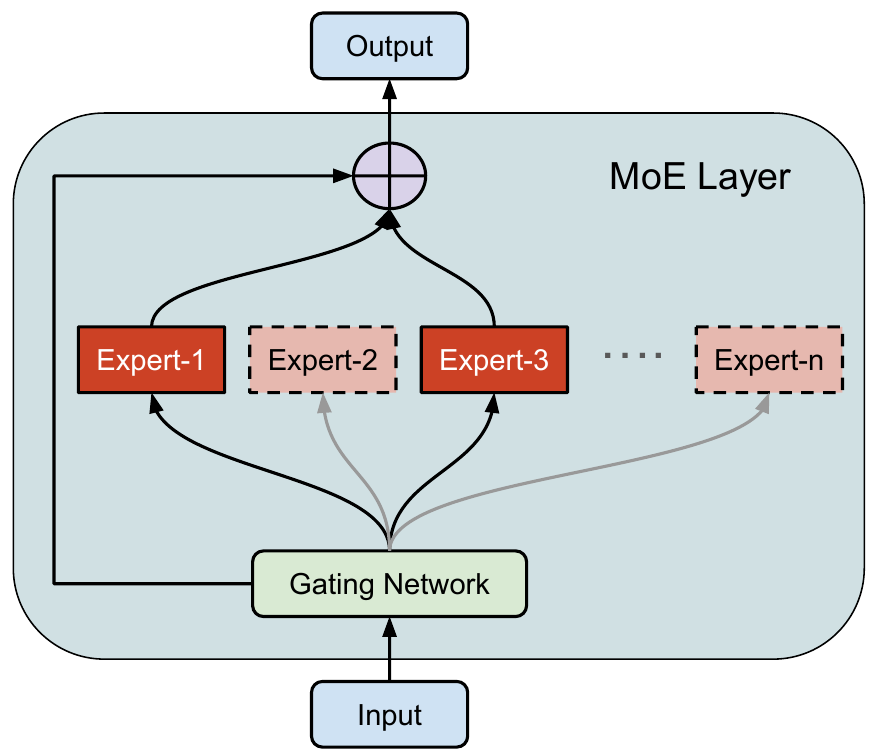}
    \caption{Mixture of Experts in Mixtral \cite{jiang2024mixtral}}
    \label{fig:Mixtral_MoE}
\end{figure}

\subsection{Transformer Modules}
\label{appendix:module}

\subsubsection{Multi-Head Self-Attention (MHSA)} In a Multi-Head Self-Attention module, each attention head computes its own unique set of query, key, and value vectors. This allows the model to attend to different subspaces of the input representation in parallel. MHSA offers the best performance but is computationally expensive and memory-intensive, especially for large models.

\begin{figure}
\centering
\begin{minipage}{.4\linewidth}
\centering
\subfloat[MHSA]{\label{fig:mhsa}\includegraphics[scale=.4]{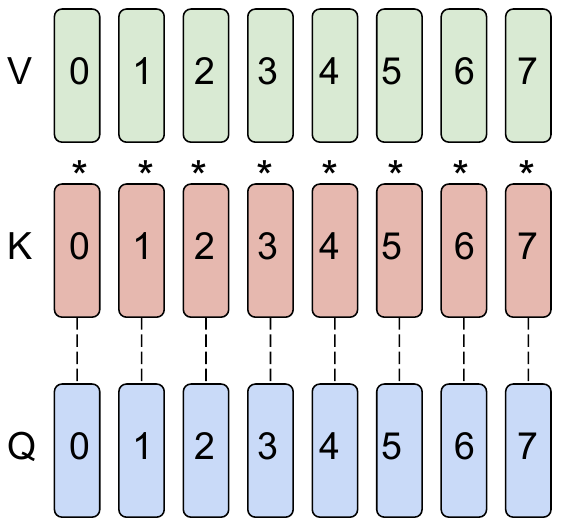}}
\end{minipage}
\quad
\begin{minipage}{.4\linewidth}
\centering
\subfloat[GQA]{\label{fig:gqa}\includegraphics[scale=.4]{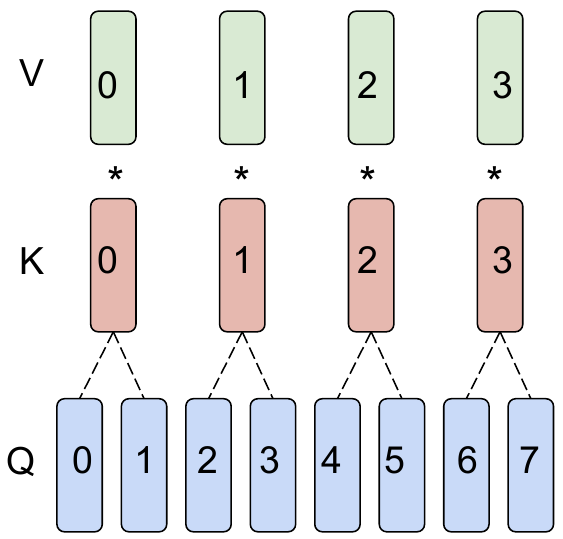}}
\end{minipage}\par\medskip
\caption{Types of Self Attention Methods}
\label{fig:attention_methods}
\end{figure}

\subsubsection{Group Query Attention (GQA)}

Grouped Query Attention (GQA) divides query heads into multiple groups where each group shares a single key head and value head, as depicted in Figure \ref{fig:gqa}. It has reduced number of parameters compared to MHSA by sharing the key and value heads.

\subsection{LLM Series} \label{appendix:llms}

\subsubsection{LLaMA Series} LLaMA-2 and LLaMA-3 are LLMs developed by Meta. LLaMA-3 represents a significant advancement over its predecessor, with improvements in several key areas. LLaMA-3 was pre-trained on over 15 trillion tokens, a dataset seven times larger than LLaMA-2's, including four times more code and broader language coverage. LLaMA-3 utilizes OpenAI's Tiktoken for tokenization, replacing LLaMA-2's SentencePiece tokenizer. LLaMA-3, trained on a 24,000 GPU cluster, is available in 8B and 70B parameter sizes, while LLaMA-2 comes in 7B, 13B and 70B sizes. Notable improvements include stronger reasoning abilities, better code generation, and improved instruction following. Additionally, LLaMA-3 doubles the context window from LLaMA-2's 4K tokens to 8K tokens, allowing for more comprehensive information processing.

\subsubsection{Mistral and Mixtral} Mistral and Mixtral are LLMs developed by Mistral AI for complex NLP tasks. Mistral-7B features sliding window attention, GQA, and a byte-fallback byte pair encoding tokenizer, enabling efficient handling of long sequences while maintaining high performance. Its architecture includes an 8k context length with a theoretical attention span of 128K tokens and improved robustness via its tokenizer. Mixtral, an evolution of Mistral, introduces a sparse mixture of experts (SMoE) model. The Mixtral-8x7B model contains 45 billion parameters and outperforms its predecessors on various benchmarks while offering 6x faster inference. It employs eight experts per MLP and utilizes Flash Attention 2 for optimized attention mechanisms.

\subsubsection{Qwen 2 Series} Alibaba Cloud's Qwen 2 series represents a significant advancement, with the Qwen2-7B and Qwen2-72B models showcasing exceptional capabilities. The Qwen2-7B model, with 7.07 billion parameters (5.98 billion non-embedding), is designed for robust language tasks, while the larger Qwen2-72B model, boasting 72.71 billion parameters (70.21 billion non-embedding), is tailored for highly complex tasks and extensive datasets. Both models utilize GQA and support an impressive context length of 128K tokens, excelling in handling long texts. These models demonstrate superior performance in coding, mathematics, and multilingual proficiency, surpassing existing open-source models. The Qwen2 series' ability to handle extended context lengths, with all Instruct models trained on 32K-token contexts and capable of even longer extensions. 

Table \ref{table:model_Summary} summarizes the neural architecture configurations, which include the number of layers, hidden size, attention type, number of attention and KV heads, FFN type, number of FFN experts, FFN intermediate size and maximum sequence length and vocabulary size.

\begin{table*}
\centering

\begin{tabular}{|c|c|c|c|c|c|c|c|c|c|c|c|c|}
\hline
Models & 
\begin{tabular}[c]{@{}c@{}}\#Hidden\\Layers\end{tabular} &
\begin{tabular}[c]{@{}c@{}}Hidden\\Size\end{tabular} &
\begin{tabular}[c]{@{}c@{}}Attention\\Type\end{tabular} &
\begin{tabular}[c]{@{}c@{}}\#Attention\\Heads\end{tabular} &
\begin{tabular}[c]{@{}c@{}}\#KV\\Heads\end{tabular} &
\begin{tabular}[c]{@{}c@{}}FFN\\Type\end{tabular} &
\begin{tabular}[c]{@{}c@{}}\#FFN\\Experts\end{tabular} &
\begin{tabular}[c]{@{}c@{}}FFN\\Intermediate\\Size\end{tabular} &
\begin{tabular}[c]{@{}c@{}}Max\\Sequence\\Length\end{tabular} &
\begin{tabular}[c]{@{}c@{}}Vocab\\Size\end{tabular} & 
\begin{tabular}[c]{@{}c@{}}HF\\Repo\end{tabular} 
\\ \hline
LLaMA-2-7B     & 32 & 4096 & MHSA & 32 & 32  & Dense & 1  &  11008 & 4096   & 32000  & \cite{llama2_7b} \\ \hline
LLaMA-3-8B     & 32 & 4096 & GQA  & 32 & 8   & Dense & 1  &  14336 & 8192   & 128256 & \cite{llama3_8b} \\ \hline
Mistral-7B     & 32 & 4096 & GQA  & 32 & 8   & Dense & 1  &  14336 & 32768  & 32000  & \cite{mistral_7b} \\ \hline
Qwen-2-7B      & 28 & 3584 & GQA  & 28 & 4   & Dense & 1  &  18944 & 131072 & 152064 & \cite{Qwen2_7b} \\ \hline
LLaMA-2-70B    & 80 & 8192 & GQA  & 64 & 8   & Dense & 1  &  28672 & 4096   & 32000  & \cite{llama2_70b} \\ \hline
LLaMA-3-70B    & 80 & 8192 & GQA  & 64 & 8   & Dense & 1  &  28672 & 8192   & 128256 & \cite{llama3_70b} \\ \hline
Qwen-2-72B     & 80 & 8192 & GQA  & 64 & 8   & Dense & 1  &  29568 & 131072 & 152064 & \cite{Qwen2_72b} \\ \hline
Mixtral-8x7B   & 32 & 4096 & GQA  & 32 & 8   & MoE   & 8  &  14336 & 32768  & 32000  & \cite{mixtral_8_7b} \\ \hline
\end{tabular}
\caption{LLaMA Model Family Summary}
\label{table:model_Summary}
\end{table*}

\begin{figure}
    \centering
    \begin{minipage}[b]{0.45\linewidth}
        \centering
        \includegraphics[width=0.6\textwidth]{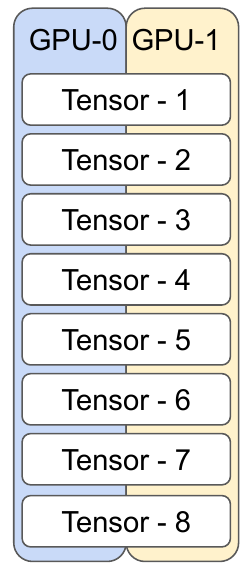}
    \end{minipage}
    \begin{minipage}[b]{0.45\linewidth}
        \centering
        \includegraphics[width=0.6\textwidth]{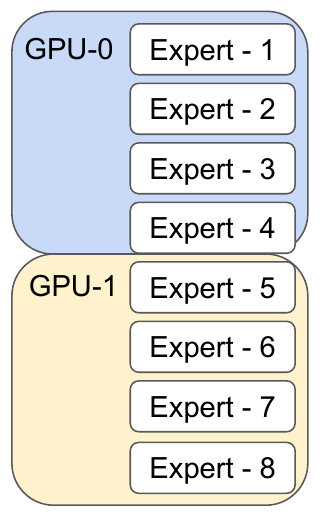}\par\vspace{5pt}
        \includegraphics[width=0.6\textwidth]{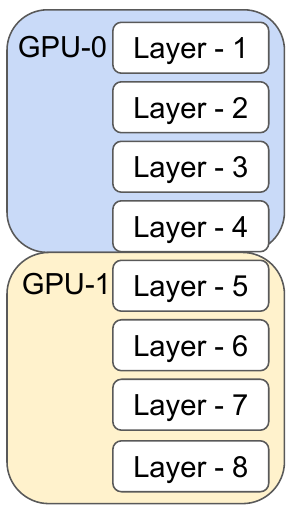}
    \end{minipage}
    \caption{LLM Parallelism Methods}
\end{figure}

\section{Hardware Platforms} \label{appendix:platforms}

The importance of AI hardware for LLMs cannot be overstated as these models continue to advance. The demand for efficient hardware to support their training and inference phases grows significantly. High-performance hardware, such as GPUs and specialized AI accelerators, enables LLMs to process vast datasets quickly and accurately, which is essential for training and deployment. The hardware platformed evaluated in this work are summarised in Table \ref{table:hwoverview}

\subsubsection{Nvidia H100 GPU} 

NVIDIA H100 GPU, which powers Meta's Datacenter for GenAI \cite{meta_datacenter}, introduces significant AI and HPC workload advancements. It is built on TSMC's 4N process with 80 billion transistors and features a dedicated Transformer Engine optimized for trillion-parameter LLMs. H100 utilizes a mix of FP8 and 16-bit calculations to achieve up to 9x faster AI training and 30x faster inference than its predecessors. The H100 also boasts fourth-generation Tensor Cores that are up to 6x faster, a 3x increase in FP64 and FP32 processing rates, and new DPX instructions that accelerate dynamic programming algorithms by up to 7x. It includes 188GB of HBM3 memory, offering nearly twice the bandwidth of the previous generation, and a 50MB L2 cache to enhance data access efficiency. The PCIe-based H100 NVL with NVLink bridge enables seamless scaling across datacenters, while the InfiniBand interconnect further boosts performance. The GPU also includes second-generation Multi-Instance GPU (MIG) technology for better resource utilization.

\subsubsection{Nvidia GH100 GPU} The NVIDIA GH200 GPU, part of the Grace Hopper Superchip architecture, combines the Hopper H100 Tensor Core GPU with the Grace CPU to deliver exceptional performance for AI and HPC. This design leverages Nvidia's ultra-fast chip-to-chip interconnect with 900GB/s bandwidth, 7x faster than PCIe Gen5. The GH200 architecture supports up to 30x higher aggregate bandwidth than today's fastest servers. The Grace CPU, featuring 72 Arm Neoverse V2 cores, provides leading per-thread performance and energy efficiency, with up to 480GB of LPDDR5X memory delivering 500GB/s of bandwidth per CPU. The GH200 NVL32 enables all GPU threads in the NVLink-connected domain to address up to 19.5TB of memory with 900GB/s bandwidth per superchip and up to 14.4TB/s bisection bandwidth in a 32 GPU system, making it ideal for large-scale AI training and HPC workloads. This platform, supported by NVIDIA MGX with GH200 and InfiniBand or Ethernet, is optimized for scale-out machine learning and HPC tasks.


\subsubsection{Nvidia A100 GPU}
The NVIDIA A100 GPU is designed for HPC AI workloads. Each A100 GPU features 6912 CUDA cores and 432 Tensor cores, supporting FP32, FP16, BF16, and Int8 precisions. It has a maximum thermal design power of 600 watts. The GPUs have 40 GiB of HBM2, offering a memory bandwidth of 6.4 TB/s. In our A100 setup, we have four A100 GPUs on each conbcedt connected via NVLink, which facilitates high-bandwidth and low-latency communication between GPUs. The host consists of a 2.8 GHz AMD EPYC Milan 7543P 32-core CPU and 512 GB of DDR4 RAM. 

\subsubsection{AMD MI250 GPU} The AMD Instinct MI250 is a GPU accelerator based on the CDNA2 architecture, manufactured using TSMC's 6nm FinFET process. It features 13,312 stream processors across 208 compute units, with a peak engine clock of 1700 MHz. The MI250 delivers impressive performance across various precision levels, including 362.1 TFLOPs for FP16, INT4, INT8, and bfloat16 operations and 90.5 TFLOPs for FP32 and FP64 matrix operations. It consists of 128 GB of HBM2e memory with an 8192-bit interface, a 1.6 GHz memory clock, and a peak memory bandwidth of 3.2 TB/s. The GPU supports PCIe 4.0 x16 and includes 8 Infinity Fabric links with 100 GB/s peak bandwidth per link with a thermal design power of 500W (560W peak). 

\subsubsection{AMD MI300X GPU} The AMD Instinct MI300X is a GPU accelerator based on the CDNA 3 architecture, manufactured using TSMC’s 5nm process. It features 19,456 shading units and 304 compute units, with a base clock of 1000 MHz that can boost up to 2100 MHz. The MI300X delivers exceptional performance across various precision levels, including 20.9 PFLOPs for FP8 operations and 5.2 PFLOPs for TF32 operations with structured sparsity. It consists of 192 GB of HBM3 memory with an 8192-bit interface, a memory clock of 2525 MHz, and a peak memory bandwidth of 5.3 TB/s. The GPU supports PCIe Gen 5 x16 and includes multiple Infinity Fabric links with a peak bandwidth of 128 GB/s per link, with a thermal design power of 750W.

\subsubsection{SambaNova SN40L} The SambaNova SN40L \cite{prabhakar2024sambanova, emani2021accelerating} Reconfigurable Dataflow Unit (RDU) is a commercial dataflow accelerator designed for enterprise inference and training applications. It features a novel three-tier memory system with 520 MiB of on-chip SRAM, 64 GiB of on-package HBM, and up to 1.5 TiB of off-package DDR DRAM, interconnected via a dedicated inter-RDU network for scalability. Each SN40L socket boasts 638 BF16 TFLOPS of peak performance, utilizing 1040 distributed Pattern Compute Units (PCUs) and Pattern Memory Units (PMUs) that deliver hundreds of TBps of on-chip memory bandwidth. This architecture enables the fusion of complex operations into single kernel calls, achieving speedups of 2× to 13× on various benchmarks compared to a baseline without the need for manual kernel programming.

\subsubsection{Habana Gaudi2} Habana Gaudi2 \cite{habana2022gaudi2} is an AI processor that has a heterogeneous compute architecture on the chip - two Matrix Multiplication Engines (MMEs) and a fully programmable 24 Tensor Processor Cores (TPCs). Each Habana’s Gaudi processor (HPU) device consists of 48 MB SRAM, 96 GB of HBM2E memory divided into six segments, and 24 100 Gigabit per second RDMA NIC Ethernet. The MME computes all operations which can be converted to matrix multiplication (fully connected layers, convolutions, batched-GEMM). In contrast, the TPC is a VLIW SIMD processor tailor-made for other DL operations. Habana Gaudi2 can support vLLM, Deepspeed and customized library Optimum Habana.

\begin{table*}[htbp]
\begin{center}
    \caption{Features of evaluated AI accelerators}     \label{table:hwoverview}
\footnotesize
\centering
    \begin{NiceTabular}{ p{1.5cm} | p{1.75cm} | p{1.75cm} | p{1.75cm} | p{1.75cm} | p{1.75cm} | p{1.75cm} | p{1.75cm} }
    \CodeBefore
        \rowcolor{gray!20}{1}
        \rowcolors{2}{gray!15}{white}
    \Body
        \toprule
        \footnotesize{\textbf{Feature}} &
        \footnotesize{\textbf{Nvidia A100}} &
        \footnotesize{\textbf{Nvidia H100}}  &
        \footnotesize{\textbf{Nvidia GH200}} &
        \footnotesize{\textbf{AMD MI250}} &  
        \footnotesize{\textbf{AMD MI300X}} &
        \footnotesize{\textbf{Habana Gaudi2}} &
        \footnotesize{\textbf{SambaNova SN40L}} \\
        \toprule
        \# Devices & 
        $4$& 
        $4$&
        $1$&
        $4$& 
        $8$& 
        $8$&
        $8$\\
        
        Memory \footnotesize{($/ \mathrm{node}$)} & 
        160 GB &
        320 GB & 
        96 GB &
        512 GB &
        1536GB &
        768 GB &
        512 GB \\

        Memory \footnotesize{($/ \mathrm{device}$)} & 
        40 GB &
        80 GB & 
        96 GB &
        128 GB &
        192GB &
        96 GB &
        64 GB \\

        Interconnect & 
        NVLink & 
        NVLink & 
        N/A & 
        Infinity Fabric & 
        Infinity Fabric & 
        RoCE V2  &
        PCIe Inter-RDU network \\    
        
        Inference Framework & 
        TensorRT-LLM, vLLM, llama.cpp, Deepspeed-MII  &
        TensorRT-LLM, vLLM, llama.cpp, Deepspeed-MII  &
        TensorRT-LLM, vLLM, llama.cpp, Deepspeed-MII  &
        vLLM, llama.cpp, Deepspeed-MII &
        vLLM, llama.cpp, Deepspeed-MII &
        vLLM, Deepspeed &        
        SambaFlow\textsuperscript{TM} \\

        Precision Support & 
        FP32, FP16, BF16, INT8, INT4, INT1 & 
        TF32, FP32, FP16, BF16, FP8, INT8, INT4, INT1  &
        TF32, FP32, FP16, BF16, FP8, INT8, INT4, INT1  &
        FP32, FP16, BF16, INT8 & 
        FP32, FP16, BF16, FP8, INT8 & 
        BF16, FP16, FP8 & 
        FP32, BF16, INT32, INT16, INT8 \\
        
        Compute Units \footnotesize{($/ \mathrm{device}$)}& 
        6912 Cuda Cores, 432 Tensor Cores & 
        16896 Cuda Cores, 456 Tensor Cores &
        16896 Cuda Cores, 456 Tensor Cores &
        208 Compute Units &
        304 Compute Units, 1216 Matrix Cores  &
        24 TPC + 2 MME & 
        1040 PCU and PMU \\

      \bottomrule
    \end{NiceTabular}
\end{center}
 \vspace{-8mm}
\end{table*}

\section{LLM Inference Frameworks}
\label{appendix:infr_frameworks}

\subsubsection{TensorRT-LLM} 
TensorRT-LLM is a powerful toolkit that provides an intuitive Python API for defining LLMs and building optimized TensorRT engines for efficient inference on NVIDIA GPUs. It incorporates SOTA optimizations, including kernel fusion, padded and packed tensors, quantization, and runtime optimizations like C++ implementations, KV caching, continuous in-flight batching, and paged attention. The library offers components to create both Python and C++ runtimes for executing the optimized TensorRT engines, enabling users to harness the full potential of LLMs across various configurations, from single GPUs to multi-node setups with multiple GPUs using Tensor, Pipeline and Expert Parallelisms. TensorRT-LLM supports a wide range of popular LLM architectures and includes features such as beam search, extensive sampling functionalities, and integration with the NVIDIA Triton Inference server for production-quality deployment for applications from real-time chatbots to complex text analysis. 

\subsubsection{vLLM}  
The vLLM framework \cite{kwon2023efficient} is an open-source, high-performance solution developed initially at UC Berkeley. Now, it is a community project designed to serve and optimize the deployment of LLMs. At its core, it utilizes an optimized attention algorithm called PagedAttention, which dynamically allocates GPU memory for actual decoding lengths, significantly reducing memory consumption and increasing throughput. vLLM supports many popular and SOTA LLMs and incorporates various modern LLM acceleration techniques such as speculative decoding, chunked prefill, flash attention, HIP and CUDA graphs, tensor parallel multi-GPU, and quantization methods such as GPTQ \cite{frantar2022gptq}, AWQ \cite{lin2023awq}, SqueezeLLM \cite{kim2023squeezellm}, FP8 KV Cache. vLLM's continuous batching feature allows it to process multiple requests simultaneously to tackle heavy query loads effectively. The framework offers a simple Python API for offline inference and an OpenAI API-compatible server for online serving. 

\subsubsection{Deepspeed} DeepSpeed is an open-source deep learning optimization library developed by Microsoft, primarily designed to enhance the efficiency and effectiveness of distributed training and inference for large-scale models. It incorporates several optimizations, such as Zero Redundancy Optimizer (ZeRO), to distribute model states across GPUs to reduce communication and 3D Parallelism to combine multiple parallelisms. It can accelerate LLMs on different hardware platforms, such as Nvidia GPUs.  


\subsubsection{Deepspeed-MII}  
DeepSpeed-MII is an open-source library from DeepSpeed that develops low-latency, low-cost solutions for LLM inference. This framework leverages extensive optimizations from DeepSpeed-Inference, such as deepfusion for transformers, automated tensor-slicing for multi-GPU inference, compiler optimizations via TorchScript and nvFuser, and on-the-fly quantization with ZeroQuant. MII also features blocked KV-caching, continuous batching, Dynamic SplitFuse, tensor parallelism, and high-performance CUDA kernels to support fast, high-throughput text generation. 

\subsubsection{llama.cpp} 
llama.cpp is an open-source, high-performance portable inference framework for LLMs written in C/C++, a viable alternative to heavyweight frameworks. It stands out for its ability to run models efficiently on consumer-grade hardware, making LLM inference accessible to users without specialized equipment. The framework employs advanced optimization techniques, including quantization and efficient memory mapping, to significantly reduce the memory footprint of LLMs without substantial performance degradation. llama.cpp's lightweight design ensures fast responses and broad compatibility across various platforms, from CPUs to GPUs. It supports multiple hardware acceleration options, including CUDA for NVIDIA GPUs, METAL for Apple M1/M2 chips, and CLBLAST for AMD/Intel GPUs. The project's focus on efficiency, portability, and customization has made it a valuable tool for researchers and developers working with LLMs in resource-constrained environments or exploring AI capabilities on common hardware. It has bindings across several programming languages. Notable ones include llama-cpp-python, a Python interface on top of C++.  

Table \ref{table:framework_summary} summarizes the different hardware platforms and inference frameworks we utilized in our study.

\begin{table}[H]
\centering

\resizebox{\linewidth}{!}{
\begin{tabular}{|c|c|c|c|c|c|c|}
\toprule
\multicolumn{1}{|c|}{\textbf{Framework}} & \multicolumn{1}{c|}{\textbf{\begin{tabular}[c]{@{}c@{}}NVIDIA\\ A100\end{tabular}}} & \multicolumn{1}{c|}{\textbf{\begin{tabular}[c]{@{}c@{}}NVIDIA\\ H100\end{tabular}}} & \multicolumn{1}{c|}{\textbf{\begin{tabular}[c]{@{}c@{}}NVIDIA\\ GH200\end{tabular}}} & \multicolumn{1}{c|}{\textbf{\begin{tabular}[c]{@{}c@{}}AMD\\ MI250\end{tabular}}} & \multicolumn{1}{c|}{\textbf{\begin{tabular}[c]{@{}c@{}}Habana\\ Gaudi2\end{tabular}}} \\ \midrule

vLLM & Yes & Yes & Yes & Yes & Yes \\
llama.cpp & Yes & Yes & Yes & Yes & N/A \\
TensorRT-LLM & Yes & Yes & Yes & N/A & N/A \\
Deepspeed-MII & Yes & No & No & No & Yes \\ \bottomrule

\end{tabular}}
    \caption{Summary of Inference Frameworks Evaluated}
    \label{table:framework_summary}
\end{table}

\section{Miscellaneous}

\subsection{Perpleixty}

We evaluate the LLMs on LongBench \cite{bai2023longbench}, an open-source benchmark consisting of the following datasets: HotpotQA \cite{yang2018hotpotqa}, 2wikimqa \cite{ho2020constructing}, musique \cite{trivedi2022musique}, DuReader \cite{he2018dureader}, narrativeqa \cite{kovcisky2018narrativeqa}, qasper \cite{dasigi2021dataset}, GovReport \cite{huang2021efficient}, QMSum \cite{zhong2021qmsum}, VCSUM \cite{wu2023vcsum}, TriviaQA \cite{joshi2017triviaqa}, SAMSum \cite{gliwa2019samsum}, multi-news \cite{fabbri2019multi}, trec \cite{li2002learning}, lcc \cite{guo2023longcoder}, repobench \cite{liu2023repobench}. We combine all these datasets and evaluate models on the large unified dataset. Figure \ref{fig:7b_Perplexity_vs_Throughput_H100} compares the perplexity vs throughput of several $\sim$7B models on H100 GPU, evaluated on LongBench dataset \cite{bai2023longbench}. The models include LLaMA-2-7B, LLaMA-3-8B, Aquila-7B, Qwen1.5-7B, OPT-6.7B, LLaMA-7B, GPT-J-6B, Bloom-7.1B, and DeciLM-7B. The LLaMA-2-7B model shows best perplexity but low throughput compared to LLaMA-3-8B. In contrast, DeciLM-7B has the highest throughput with $~$5.5k tokens per second.

\begin{figure}
    \centering
        \includegraphics[width=\linewidth]{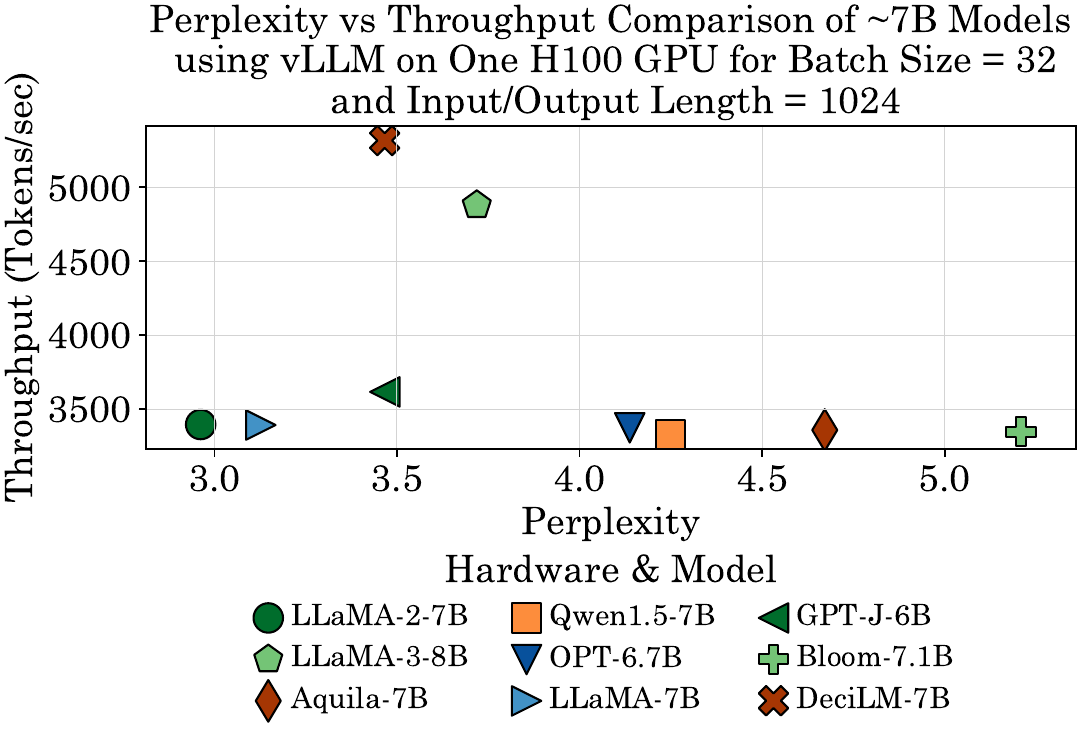}
        \caption{H100: Perplexity vs Throughput}
        \label{fig:7b_Perplexity_vs_Throughput_H100}
    \captionsetup{justification=centering}
\end{figure}

\section{Additional Results}

In this section, we provide additional benchmarking results to better understand accelerators, frameworks and models.  

\subsection{TensorRT-LLM} Figure \ref{fig:appendix_A100_7B} illustrates the performance of 7B models across different batch sizes and the number of GPUs using TensorRT-LLM. Throughput increases as batch size increases for all models and the number of GPU computing devices. LLaMA-2-7B model performance saturates with a decrease in the number of GPUs, and Mistral-7B outperforms LLaMA-3-8B across different batch sizes and number of GPUs.

\begin{figure}
    \centering
    \includegraphics[width=0.90\linewidth]{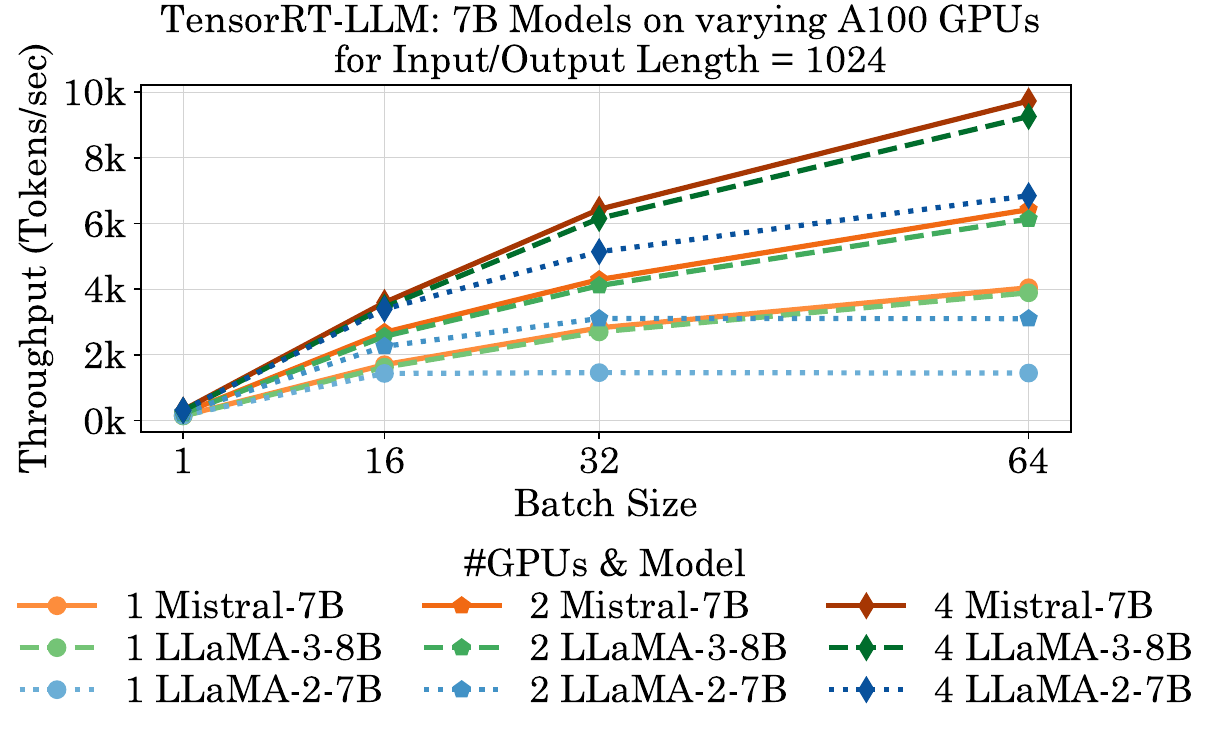}
    \captionsetup{justification=centering}
    \caption{TRT-LLM: 7B Models on 1,2 and 4 A100 GPUs}
    \label{fig:appendix_A100_7B}
\end{figure}

\subsection{vLLM}

Figure \ref{fig:appendix_vLLM_7B} illustrates the performance of 7B models across different batch sizes and the number of H100, A100 and MI250 GPUs using vLLM. The H100 systems consistently achieve higher throughput across all models and number of computing devices. Despite having 1 billion more parameters, LLaMA-3-8B outperforms Mistral-7B on H100 GPU. This shows that H100 can handle large models using vLLM than TensorRT-LLM.

\begin{figure}
    \centering
    \includegraphics[width=0.90\linewidth]{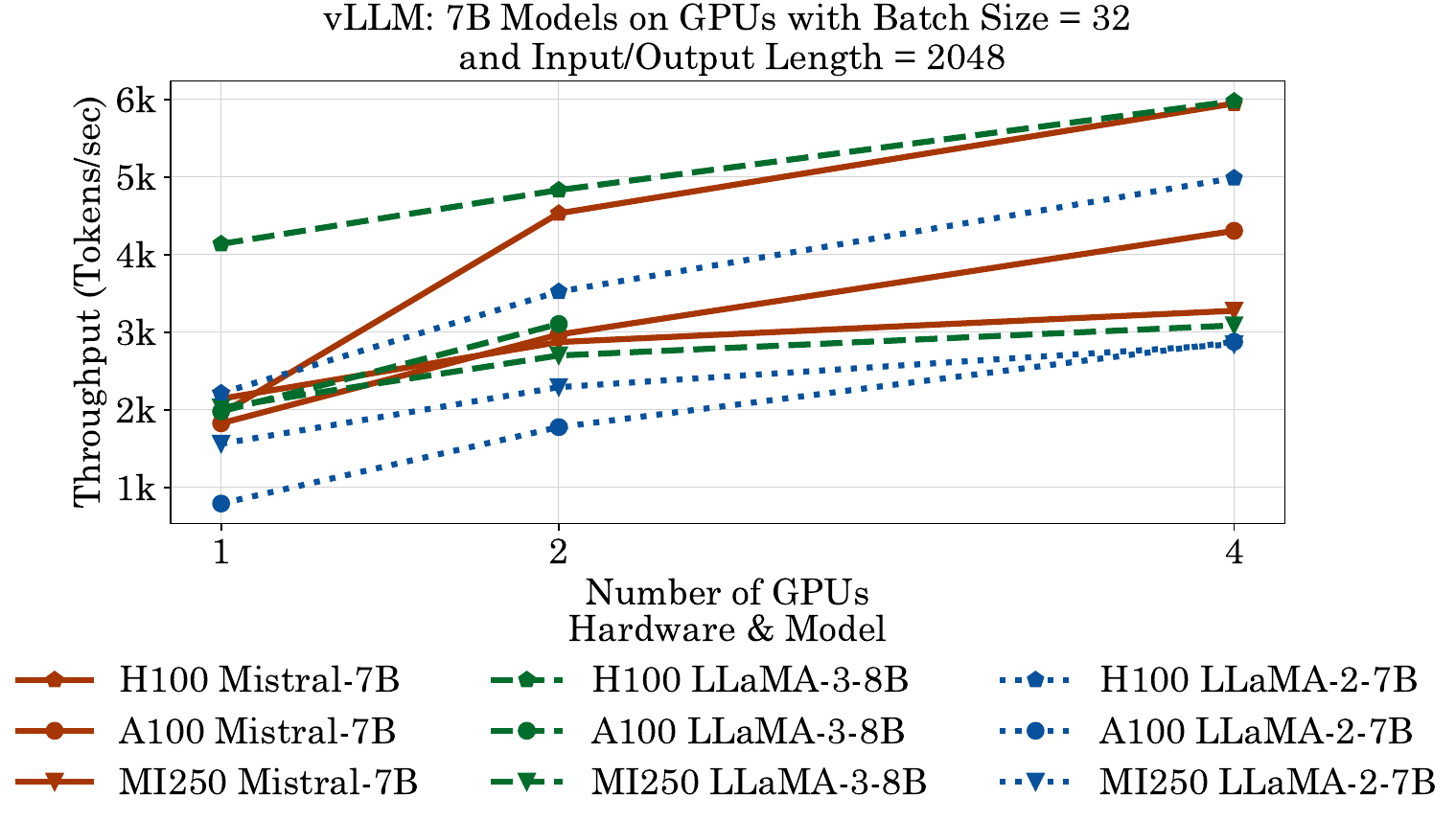}
    \captionsetup{justification=centering}
    \caption{vLLM: 7B Models on 1,2 and 4 GPUs}
    \label{fig:appendix_vLLM_7B}
\end{figure}

\subsection{llama.cpp}

Figure \ref{fig:appendix_vLLM_7B} depicts the performance of 70B models across different batch sizes on four GPUs using llama.cpp. We exclude A100 numbers from the figures as the 70B models could not fit on one A100 node which consists of 40GB on each chip. H100 GPUs perform better than MI250 GPUs, and Mixtral-8x7B outperforms LLaMA-2-70b and LLaMa-3-70b due to a sparse mixture of expert modules. 

\begin{figure}
    \centering
    \includegraphics[width=0.8\linewidth]{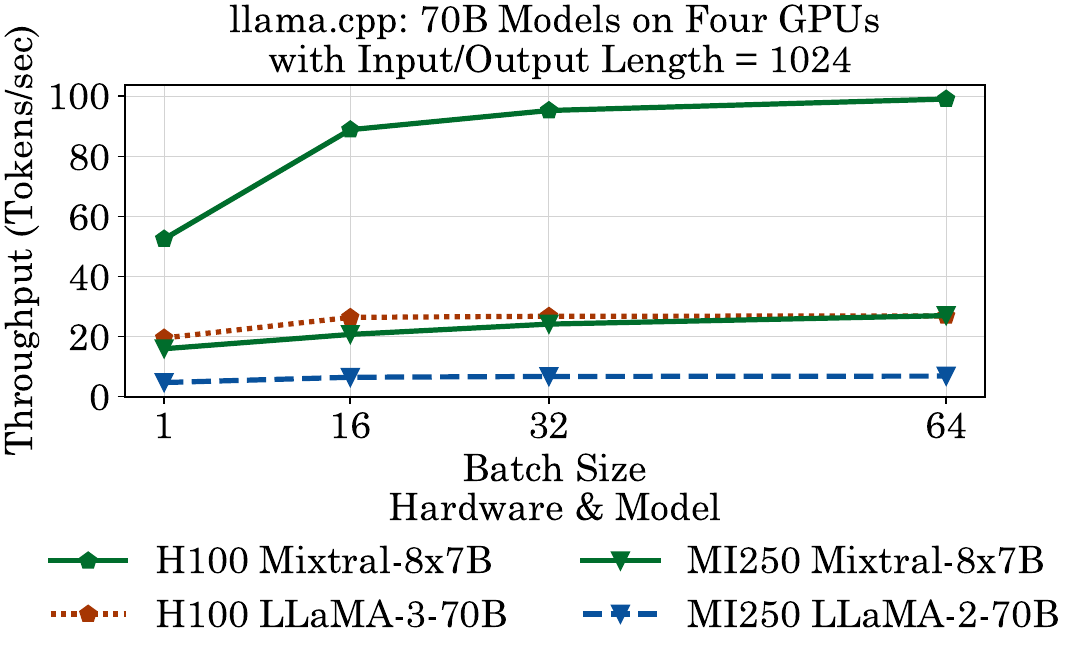}
    \captionsetup{justification=centering}
    \caption{llama.cpp: 70B Models on H100 and MI250}
    \label{fig:appendix_llama_cpp_70b}
\end{figure}

\subsection{Nvidia GPUs}

Figure \ref{fig:appendix_H100_7B} compares different frameworks and $~$7B models on H100 GPU for input and output length 1024, where Qwen2-7B with TRT-LLM attains the highest throughput and the next closet performer being Qwen2-7B with vLLM. This is due to less number of neural architecture hyperparameters for Qwen2-7B, such as numbers of layers, hidden size, and FFN dimension, compared to other models (See Table \ref{table:model_Summary}).   

\begin{figure}
    \centering
    \includegraphics[width=0.8\linewidth]{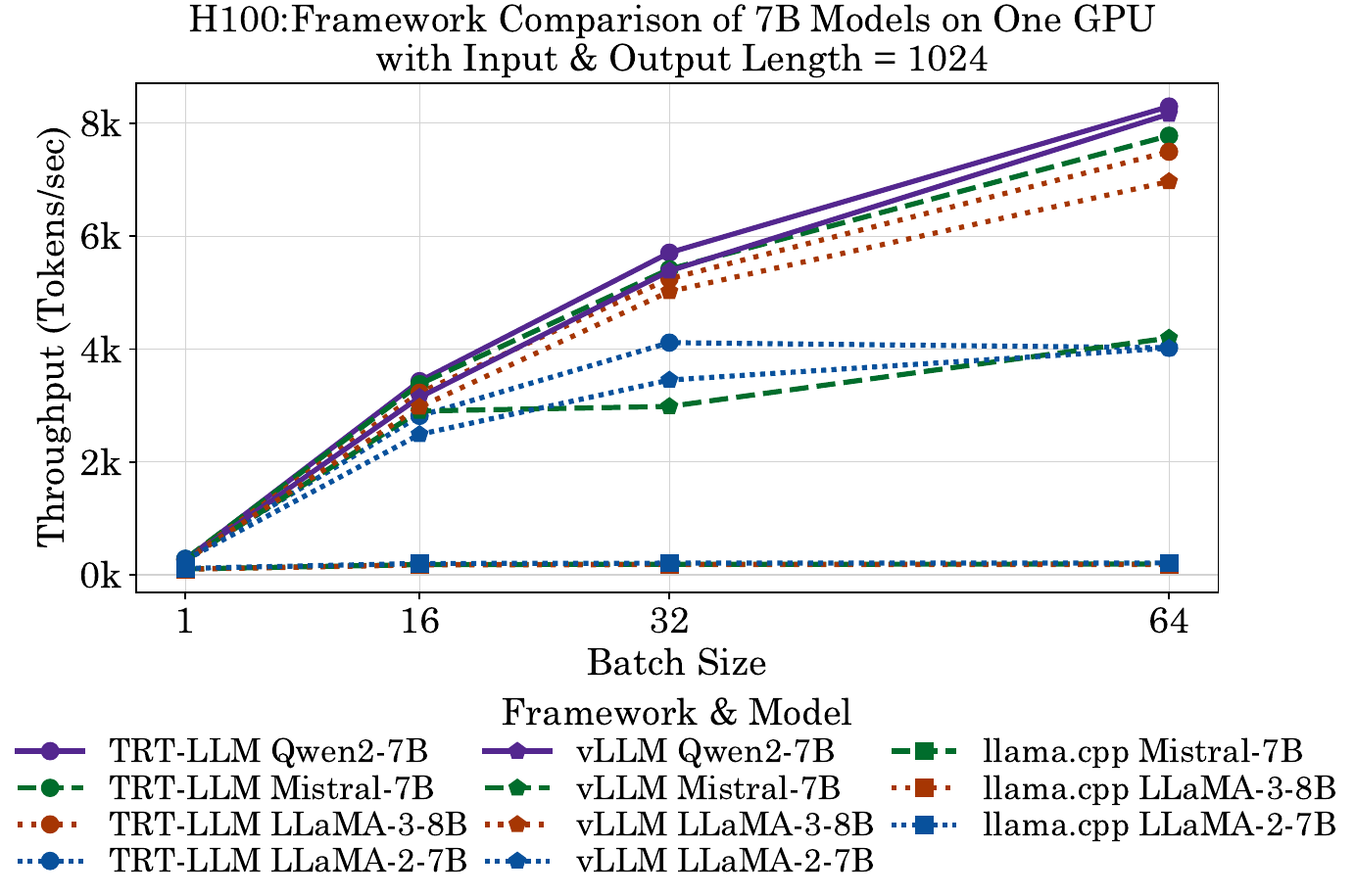}
    \captionsetup{justification=centering}
    \caption{7B Model Framework Comparison on H100}
    \label{fig:appendix_H100_7B}
\end{figure}

In Figure \ref{fig:appendix_A100_H100_70B}, we compare the execution performance of TRT-LLM and vLLM on A100 and H100 GPUs for $\sim$70B models. MoE model Mixtral outperforms 70B models by a considerable margin, whereas LLaMA-2-70B using vLLM and TRT-LLM performs slightly better than LLaMA-3-70B on A100 and H100. 

\begin{figure}
    \centering
    \includegraphics[width=0.8\linewidth]{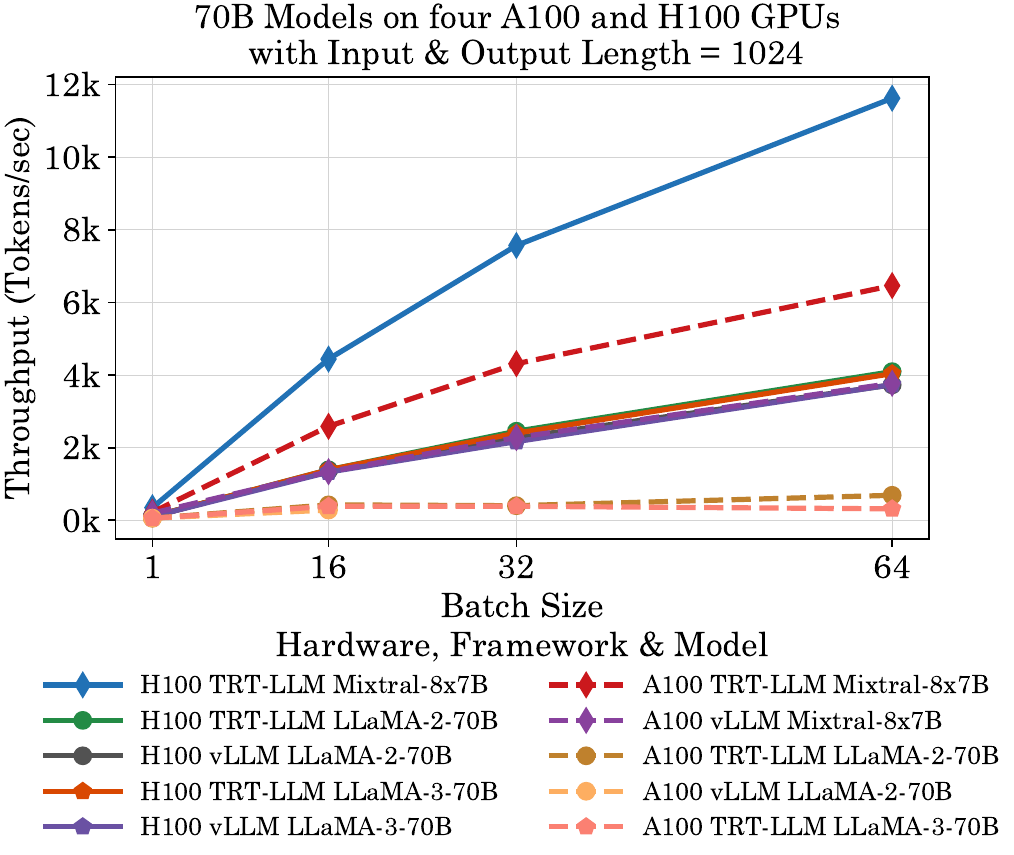}
    \captionsetup{justification=centering}
    \caption{70B models on A100 and H100}
    \label{fig:appendix_A100_H100_70B}
\end{figure}

\subsection{AMD MI250}

Figure \ref{fig:appendix_MI250_7B_vLLM} illustrates the performance of Qwen2-7B, Mistral-7B, LLaMA-3-8B, LLaMA-2-7B models on MI250 GPU using vLLM across different batch sizes. We observe that Qwen2-7B, Mistral-7B and LLaMA-38B models attain their peak performance at batch size 32 and decline for batch size 64. However, LLaMA-2-7B achieves the highest throughput than other models at batch size 64. This is contrary to other hardware, as LLaMA-2-7B with MHSA should saturate faster than models with GQA. Within batch size 32, Qwen2-7B outperforms Mistral-7B and Mistral-7B slightly better than LLaMA-3-8B due to its relatively smaller vocab size. 

\begin{figure}
    \centering
    \includegraphics[width=0.8\linewidth]{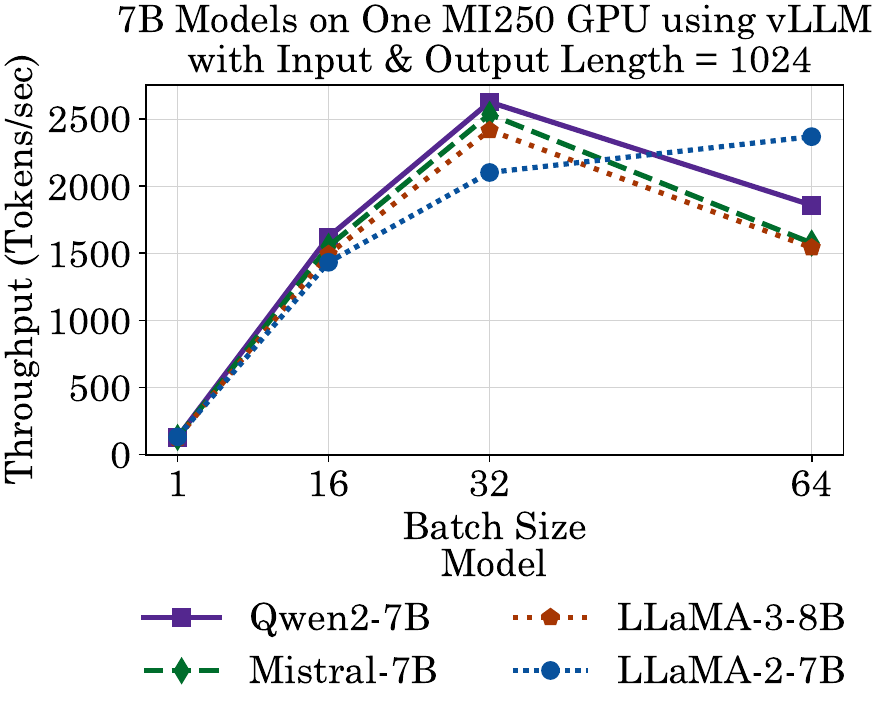}
    \captionsetup{justification=centering}
    \caption{MI250: vLLM on 7B Models}
    \label{fig:appendix_MI250_7B_vLLM}
\end{figure}

Figure \ref{fig:appendix_MI250_7B_llamacpp} illustrates the performance of Qwen2-7B, Mistral-7B, LLaMA-3-8B, LLaMA-2-7B models on MI250 GPU using llama.cpp across different batch sizes. LLaMA-2-7B using llama.cpp on MI250 attains the best performance across all batch sizes compared to other models. This concludes that llama.cpp cannot better utilize GQA as models with GQA lag behind MHSA. Qwen2-7B, the model with the best performance using vLLM has the least performance using llama.cpp on MI250 GPU. Figure \ref{fig:appendix_MI250_70B_vllm} compares several large (MoE and 70B) models on 4 MI250 GPUs using vLLM. Similar to TRT-LLM, Mixtral-8x7B attains higher performance than other models. We also observe that all models scale well with an increase in the number of GPUs.

\begin{figure}
    \centering
    \includegraphics[width=0.75\linewidth]{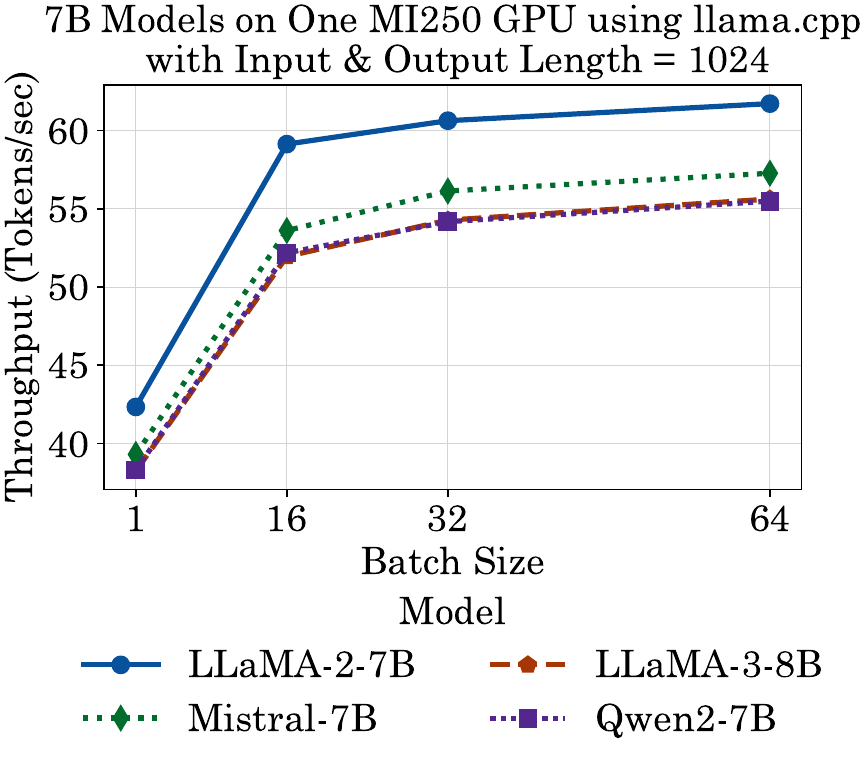}
    \captionsetup{justification=centering}
    \caption{MI250: llama.cpp on 7B Models}
    \label{fig:appendix_MI250_7B_llamacpp}
\end{figure}

\begin{figure}
    \centering
    \includegraphics[width=\linewidth]{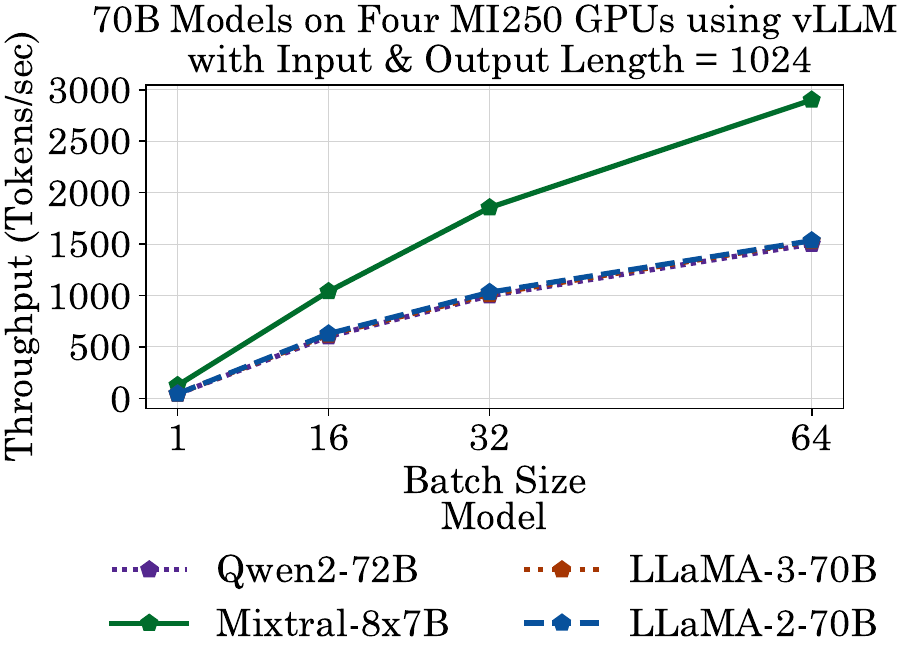}
    \captionsetup{justification=centering}
    \caption{MI250: vLLM on 70B Models}
    \label{fig:appendix_MI250_70B_vllm}
\end{figure}

\subsection{Habana Gaudi2} Figure \ref{fig:70B_Gaudi2} compares the performance of several 70B models on Gaudi2, H100 and A100. We observe that the performance of Gaudi2 lies between H100 and A100 across all the models, while Gaudi2 outperforms A100, lagging behind H100.

\begin{figure}[H]
\centering
\includegraphics[width=\linewidth]{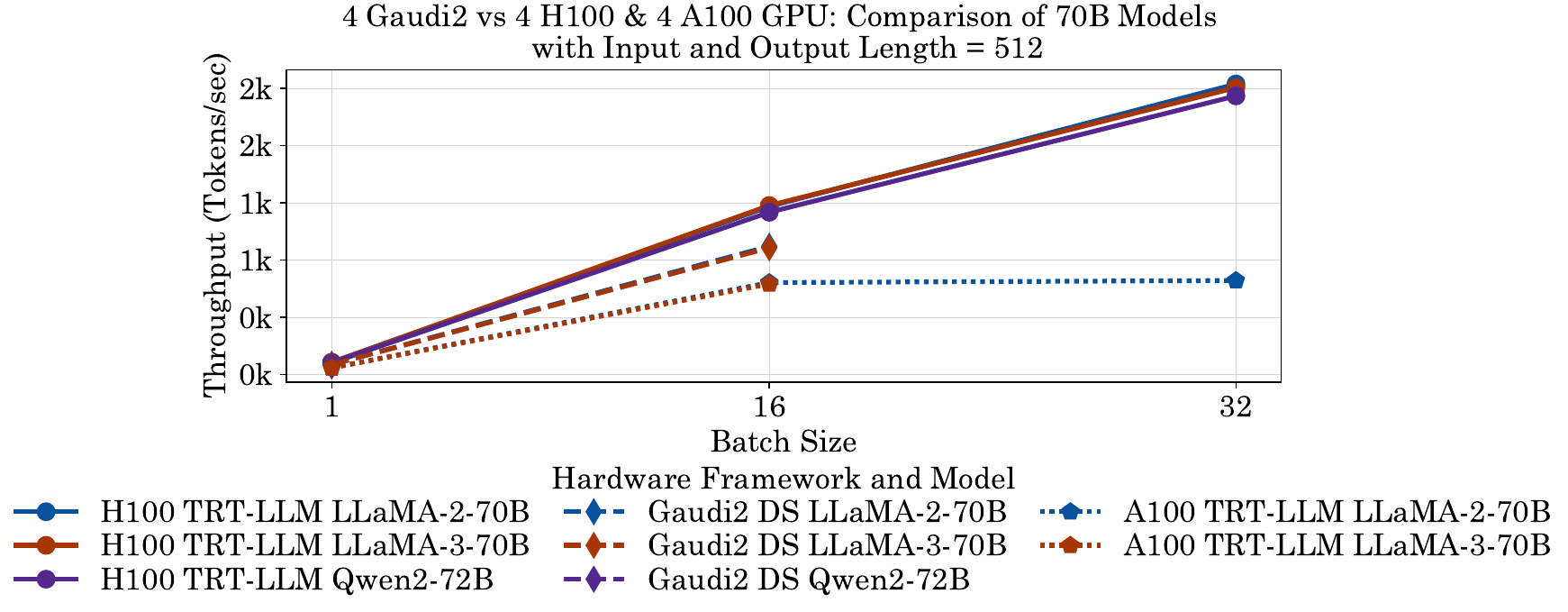}
\caption{H100 vs A100 vs Gaudi2: 70B Models}
\label{fig:70B_Gaudi2}
\captionsetup{justification=centering}
\end{figure}

\section{Artifact Description}

All the instructions to download weights, set up frameworks, run benchmarks and plot results are present in the Github repository \href{https://github.com/argonne-lcf/LLM-Inference-Bench}{https://github.com/argonne-lcf/LLM-Inference-Bench}.  
\newpage

\end{document}